%% file: main.tex
\documentclass[10pt,twocolumn,letterpaper]{article}

\usepackage{cvpr}              

\input{preamble}

\usepackage{booktabs}
\usepackage{multirow}
\usepackage{rotating}
\usepackage{adjustbox}
\def\eg{\emph{e.g}\onedot}

\def\ie{\emph{i.e}\onedot}

\usepackage{mathtools}
\usepackage{textcomp}
\usepackage{gensymb}
\usepackage{amsmath}
\usepackage{soul}
\usepackage{bbm}
\usepackage[dvipsnames]{xcolor}

\newcommand{\x}[1]{{\color{red}{placeholder}}} 

\definecolor{cvprblue}{rgb}{0.21,0.49,0.74}
\usepackage[pagebackref,breaklinks,colorlinks,citecolor=cvprblue]{hyperref}

\def\paperID{10404} 
\def\confName{CVPR}
\def\confYear{2024}

\title{EarthLoc: Astronaut Photography Localization by Indexing Earth from Space}

\author{Gabriele Berton$^{1}$
\quad
Alex Stoken$^{2}$
\quad
Barbara Caputo$^{1}$
\quad
Carlo Masone$^{1}$\\
$^{1}$Politecnico di Torino 
$^{2}$Jacobs Technology, NASA Johnson Space Center\\
{\tt\small gabriele.berton@polito.it}\\
}

\begin{document}
\maketitle
\input{sec/0_abstract}

\input{sec/1_introduction}

\input{sec/2_related_work}

\input{sec/3_dataset}

\input{sec/4_method}

\input{sec/5_experiments}

\input{sec/6_conclusions}
{
    \small
    \bibliographystyle{ieeenat_fullname}
    \bibliography{main}
}

\input{sec/X_suppl}

\end{document}

%% file: preamble.tex
\usepackage[dvipsnames]{xcolor}


%% file: sec/0_abstract.tex
\begin{abstract}
Astronaut photography, spanning six decades of human spaceflight, presents a unique Earth observations dataset with immense value for both scientific research and disaster response.
Despite their significance, accurately localizing the geographical extent of these images, which is crucial for effective utilization, poses substantial challenges.
Current, manual localization efforts are time-consuming, motivating the need for automated solutions. We propose a novel approach – leveraging image retrieval – to address this challenge efficiently. We introduce innovative training techniques which contribute to the development of a high-performance model, EarthLoc. We develop six evaluation datasets and perform a comprehensive benchmark comparing EarthLoc to existing methods, showcasing its superior efficiency and accuracy. Our approach marks a significant advancement in automating the localization of astronaut photography, which will help bridge a critical gap in Earth observations data.
Code and datasets are available at 
{\small{\url{https://github.com/gmberton/EarthLoc}}}.
\end{abstract}

%% file: sec/1_introduction.tex
\input{figures/teaser_task_overview}

\section{Introduction}
\label{sec:introduction}
Astronaut photography of Earth is a unique remote sensing dataset that spans 60 years of human spaceflight, offering a rare perspective on our planet to the public and valuable data to Earth and atmospheric science researchers.
This dataset contains over 4.5 million images and is growing by the tens of thousands per month, as astronauts are continually tasked with taking new photographs that enable scientific research as well as assist in natural disaster response efforts in the wake of events like floods and wildfires.
To effectively use these images, the
geographical area depicted in them needs to be identified. Unfortunately, this task - Astronaut Photography Localization (\textbf{APL}) - is very challenging.
For each photo, only a coarse estimate of location is known, given by the point on Earth directly under the International Space Station (ISS) at the time the photo is taken. This point -- called the \textbf{nadir} -- can be easily computed using the image's timestamp and the ISS's orbit path.
However, two images taken with the same nadir can be thousands of kilometers apart, as even a slight inclination of the astronaut's hand-held camera can move the image's location hundreds of kilometers in any direction, as depicted in \cref{fig:teaser_task_overview}.
Localization must thus be performed over a wide area, and is additionally complicated by 
(i) astronauts using hand-held cameras and a variety of zoom lenses, (ii) the large, 2500 kilometer (km) visibility range in all directions, (iii) most photographs being oblique, and (iv) the Earth's appearance changing over time, mostly due to weather or seasonal variations.
Despite these challenges, the high value of properly localized imagery for Earth science research and disaster response has lead to more than 300,000~\cite{fisher_issrd} astronaut photos being manually localized by identifying the geographic coordinates of the photo's center pixel. This process can take minutes to hours per photo for experts in NASA's Earth Science and Remote Sensing Unit (ESRS) and citizen scientists through ESRS's Image Detective\footnote{\url{https://eol.jsc.nasa.gov/BeyondThePhotography/ImageDetective/}} program.
The resulting geo-located images have lead to a multitude of peer reviewed papers, technical reports, and articles\footnote{\url{https://eol.jsc.nasa.gov/AboutCEO/PubList.htm}} about light pollution and urban planning\cite{env_risks_artificial_light_eu, environ_impacts_artificial_light, spectrometry_urban_lightscape, nightlight_behavior}, atmospheric phenomena \cite{sprites, TLEs}, changes in land usage and glacial extent \cite{astro_photos_climate_patterns}, and other Earth science topics \cite{wilkinson_gunnell_2023, DIEGO_thermal_mission, sunglint}.
Most importantly, astronaut photography has a fast response time, crucial for disaster management - astronauts are alerted of a disaster, provide photos to ground crew which localizes them, and send them to first responders\footnote{\url{https://storymaps.arcgis.com/stories/947eb734e811465cb0425947b16b62b3}}~\cite{IDC_Stefanov}.
For example, in 2013 this protocol was activated in response to Cyclone Haiyan in the Philippines, wildfires in Australia, flooding in China, Russia, Pakistan, and the USA, and multiple other events.~\cite{IDC_Stefanov}.

Even with the abundance and importance of satellite imagery in modern applications, astronaut photography fills an unserved gap among other remotely sensed data. 
Unlike satellites that nominally take top-down imagery in similar illumination conditions at fixed temporal intervals, astronaut photographs can show topography via oblique views from multiple orientations, are taken in various lighting conditions (including nighttime), and vary in focal length to show detail at different resolutions. These qualities produce a complementary data product that would be difficult or impossible to gather from traditional satellites. Additionally, having a human in the loop for data collection allows for real-time response to natural disasters as well as a natural sense to avoid clouds and other obstructions.

Given the uniqueness and importance of astronaut photography, and the large amount of time spent by human experts on geo-localizing them, researchers have been exploring solutions to automate the task through computer vision methods.
Previous works have shown promising results using a pairwise matching setup that iteratively compares an astronaut photo to satellite imagery \cite{Stoken_2023_CVPR}. Yet, such methods suffer from high latency, searching multiple directions sequentially and using compute-heavy, dense correspondences to determine matching. The time required to find the location of all 4.5 million images using these methods is estimated to be over 20 compute years \cite{Stoken_2023_CVPR}.
Furthermore, low latency is of paramount importance for photography in areas that are affected by natural disasters (\eg, hurricanes or wildfires), for which speedy localization can help real-time, on-the-ground operations.

To overcome the high latency of current methods, we propose to instead localize astronaut photography through image retrieval, by matching each astronaut photograph to a worldwide database containing satellite images of known position.
In reformulating this problem as a retrieval task, we come across a range of new challenges, from efficiently training a robust model to achieving low-latency inference, as well as defining a success metric and creating evaluation sets, both of which are crucial pieces when assessing how well a given model will perform when deployed to localize astronaut photography.

The ultimate goal of this paper is twofold:
(i) provide an efficient method to localize the archive of 4.5M queries as well as all new, incoming imagery, so that researchers studying a given geographical area can have access to a large amount of imagery spanning many years; 
and (ii) 
prove the viability of world-wide APL through image retrieval (which has \emph{never been attempted} and was \emph{believed to be infeasible}
\footnote{in [59] (Sec. 2.1) scientists at NASA claim that retrieval for APL is infeasible because \emph{we cannot precompute a database of reference features} due to their belief that the database must be query-specific. Results show that our pipeline can overcome this issue.})
in order to spark a new line of research, with direct benefit to all space-based photography and its users across the globe.

To achieve its goal, this paper makes the following contributions:
\begin{itemize}
    \item we propose approaching APL through image retrieval;
    \item we develop novel training techniques which show large quantitative improvement on the task;
    \item we provide six evaluation sets and a large benchmark of results with methods from a variety of relevant domains;
    \item we show that a model trained with these new techniques, \textbf{EarthLoc}, allows us to localize a large number of images, widely outperforming all other methods while being fast and efficient.
\end{itemize}
Finally, we note how EarthLoc is being used to localize ISS astronaut photographs, which are publicly available and conveniently searchable at 
\url{https://eol.jsc.nasa.gov/ExplorePhotos/}.

%% file: figures/teaser_task_overview.tex
\begin{figure}
    \begin{center}
    \includegraphics[width=0.85\columnwidth]{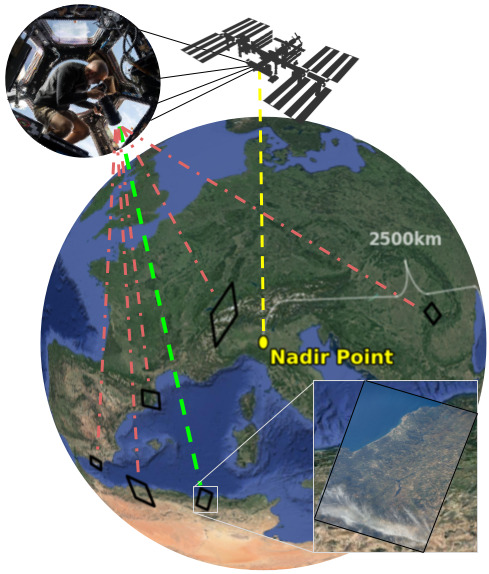}
    \end{center}
\vspace{-3mm}
    \caption{\textbf{Overview of the astronaut photography localization task.}
    Astronauts take hundreds of photos a day from the International Space Station (ISS) cupola (top-left) with hand-held cameras.
    For each image (example bottom right), the geographic location depicted is not known, and needs to be searched for across a huge area centered at the ISS's (known) \textit{nadir} point at the moment that the photo is taken.
    A simulated view of the astronaut's perspective when the ISS is above Europe is shown.
    The goal of our paper is to automate the task of localizing these images, which could be anywhere within the view.    
    In the figure's example, the photo the astronaut took is indicated by the green line and shown in inset -- other possible photo extents are in red, illustrating the wide array of potential locations to search.
    }
    \label{fig:teaser_task_overview}
\end{figure}

%% file: sec/2_related_work.tex
\section{Related Work}
\label{sec:related_work}

\paragraph{Image Retrieval}
Image retrieval involves searching a database for images similar to a query image.
Traditionally, methodologies combined hand-crafted local features like SIFT \cite{Lowe_2004_sift}, SURF \cite{Bay_2008_surf}, RootSIFT \cite{Arandjelovic_2012_rootSift} into a global embedding through means like Bag of Words \cite{Csurka_2003_bow}, Fisher Vectors \cite{Perronnin_2010_fisher} or VLAD \cite{Jegou_2011_vlad}, to allow for fast retrieval through kNN.
With the rise of deep learning, local features have been replaced by CNN-derived features \cite{Babenko_2014_neural_codes}, significantly improving retrieval performance.
Since then, most of the attention of the retrieval community has been focused on how to best aggregate feature maps and efficiently train robust neural networks.
For feature map aggregation, some of the most notable examples in literature include Max pooling \cite{Razavian_2015_mac},
Regional Max pooling (R-MAC) \cite{Tolias_2016_rmac} and Generalized Mean (GeM) \cite{Radenovic_2019_gem}.
To train these retrieval models, a number of losses have been proposed, which can be generally grouped into two categories:
stateful losses, which have weights whose size depends on the number of classes, like the 
Large-Margin Softmax \cite{Liu_2016_large_softmax},
SphereFace \cite{Liu_2017_sphereface},
CosFace \cite{Wang_2018_cosFace} and ArcFace \cite{Deng_2019_arcFace};
and their stateless counterparts, like the Contrastive and Triplet losses, and the more recent
Lifted Structure Loss \cite{Song_2016_liftedLoss},
NTXentLoss \cite{vandenOord_2018_NTXent},
Multi-Similarity Loss\cite{Wang_2019_multi_similarity_loss},
FastAP \cite{Cakir_2019_fastAP},
Supervised Contrastive \cite{Khosla_2020_SupCon} and the Circle Loss \cite{Sun_2020_CircleLoss}.

\paragraph{Geo-localization and retrieval with Aerial and Remote Sensing Imagery}
To the best of our knowledge, only two previous works have specifically addressed the localization problem for astronaut photography. Both rely on local features and compute pairwise comparisons between astronaut photography and geo-located reference imagery. Find My Astronaut Photo~\cite{Stoken_2023_CVPR} focuses on daytime imagery, presenting an evaluation of multiple methods as well as the note that pre-trained self-supervised models were not suitable for a retrieval-based approach to this task. Schwind and Storch~\cite{schwind_nighttime_georef} instead address nighttime imagery, using synthetically generated  street light maps as reference. Both works discuss the importance of proper orientation for matching and the high cost of pairwise comparisons. 

More commonly, studies of aerial imagery focus on data from unmanned aerial vehicles (UAVs) and satellite platforms. Aerial geo-localization is an important aspect of UAV navigation systems. The University-1652 dataset~\cite{zheng2020university} and associated challenge encourage work in cross-view geo-localization for drone-satellite imagery, with strong approaches modifying different parts of the general pipeline, from backbone~\cite{Zhu_2023_uav_backbone_winnerUAV_mbeg} to training setup~\cite{Deuser_2023_ogcl_uav_view}, to feature partitioning~\cite{local_pattern_network}. Most use contrastive and self-supervised losses in training. Other works address cross-view geo-localization in more extreme view point differences, like that between aerial and street views~\cite{zhu2021vigor, sample4Geo_Deuser_2023_ICCV, TransGeo_Zhu_2022_CVPR, Fervers_2023_CVPR, shi2022beyond, Shi_2022_ACCV, Shi_2023_ICCV}.

Remotely sensed data from satellites, on the other hand, often comes with reliable geo-location information from the sensor itself. Extracting features from satellite imagery is still a common task, with such features most often trained on a pretext objective before use on downstream tasks~\cite{Rolf_2021, mañas2021seasonal, ayush2021geography, satmae2022}. Occasionally, features are used for retrieval itself~\cite{hash_code_retrieval}, to identify areas of the Earth that are similar in appearance and thus might have similar properties to be studied together.
In this paper we take a separate direction from previous literature, and aim at using image retrieval techniques for APL.


\paragraph{Other localization tasks} Various other localization challenges are approached through image retrieval, including visual place recognition (VPR) and visual localization.
The former aims at coarsely localizing a given query by matching it to a database of geo-tagged photos, and is mostly studied within urban environments using street-view imagery \cite{Trivigno_2023_divideAndClassify, Barbarani_2023_local_features_benchmark}.
Performance is improved either by the use of smart aggregation (e.g. NetVLAD \cite{Arandjelovic_2018_netvlad}), attention layers (e.g. CRN \cite{Kim_2017_crn}), or through large-scale training \cite{Alibey_2022_gsvcities, Berton_2022_cosPlace, Alibey_2023_mixvpr, Leyvavallina_2021_gcl}.
More recently, the task of universal visual place recognition has been proposed by AnyLoc \cite{Keetha_2023_AnyLoc}, whose authors provide a model that performs competitively on a large range of scenarios, including aerial imagery.
The task of visual localization is focused on finding the precise camera pose of an image, and is commonly approached through feature matching techniques \cite{Sun_2021_loftr, Sarlin_2020_superglue, Barbarani_2023_local_features_benchmark, Masone_2021_survey}. It can also be used for visual place recognition \cite{Hausler_2021_patch_netvlad, Zhu_2023_r2former}, although this leads to a noticeable increase in runtime when compared to pure retrieval.

%% file: sec/3_dataset.tex
\section{Dataset}
\label{sec:dataset}

In this section we discuss the datasets that we use in this work.
Specifically, in \cref{sec:queries} we introduce the images that we need to localize, called queries;
in \cref{sec:database} we outline the collection of the database, made of satellite images (\ie, taken automatically, not with a hand-held camera) of known location;
and in \cref{sec:test_sets} we describe the creation of the evaluation datasets that we use to understand the capabilities of our models, combining queries and database in a way that will reflect the real-world use case of our method.


\subsection{Queries}
\label{sec:queries}

\input{figures/examples_queries}

Queries are taken from the Gateway to Astronaut Photography of Earth \footnote{\url{https://eol.jsc.nasa.gov/}}, a collection of over 4.5 million photos of Earth taken by astronauts on the International Space Station (\cref{fig:examples_queries}). This unique photography setting yields only coarse location information for each photograph, namely the position on Earth below the ISS (the \textbf{nadir} point) at the time the photo was taken. We seek to find the geographic area on the Earth that each photo encompasses (the photo's \textbf{location}), which can be thousands of kilometers away from the ISS nadir point.
This distance of maximum visibility (horizon) can be quickly computed as
\begin{equation}
\label{eq:d_visible}
\begin{split}
    d_{\text{visible}} &= \sqrt{2Rh + h^2} = 2436 \text{ km} 
\end{split}
\end{equation}
where $R=6317$ is Earth's radius and $h=450$ is the ISS maximum orbiting altitude.
We follow common practice and round this number to 2500 kilometers \cite{Stoken_2023_CVPR}.
\paragraph{Challenges to APL}
Some astronaut photo queries, taken with high focal length lenses, can cover only a few kilometers within the almost 20 million sq. km area visible to the astronaut taking the photo, which makes localizing the images extremely challenging.
Additional hurdles presented by the imagery acquisition process include the varying quality of the images themselves due to motion blur from the fast moving space station, and occlusion of unique, location-identifying features by cloud cover, shadow, or other portions of the ISS (\cref{fig:examples_queries}).
More queries are shown in \cref{sec:supp_further_examples} of the Supplementary.

From the 4.5 million photos available, we consider only the day-time images that have a ground truth obtained by FMAP \cite{Stoken_2023_CVPR}, and keep only those which cover areas between 5000 and 900,000 sq. km, in order to match the areas of database images (see \cref{sec:database}). This leads to 17,764 queries which we can use for validation and testing.


\subsection{Database}
\label{sec:database}

\input{figures/examples_database}

To estimate the location of the queries through image retrieval, we need a worldwide database of images, where each image is labeled with its position.
We build such a database from an open-source, composited, cloudless collection of Sentinel-2 satellite imagery\footnote{\url{https://s2maps.eu}}. This collection contains imagery of worldwide landmasses at resolutions up to 15 meters per pixel.
Some examples are shown in \cref{fig:examples_database}.

For our database we include any land area between latitudes 60° and -60° (\ie, the area traversed by the ISS). The Sentinel-2 imagery is available as map tiles, and we use tiles at zoom levels 9, 10, 11, with resolutions ranging from $\sim$300 to $\sim$75 meters per pixel\footnote{\url{https://wiki.openstreetmap.org/wiki/Zoom_levels}}, to ensure that the database encompasses the large scale range of the query images.

These parameters lead us to define roughly 175k \textbf{regions}, where each \textit{region} $\mathcal{R}$ is the projection of a square with known corner coordinates onto the Earth's surface. \textit{Regions} have areas ranging from 5000 sq. km to 900,000 sq. km.
We take partially (up to 50\%) overlapping \textit{regions} so that each query has at least one image in the database with considerable overlap (\cref{fig:examples_database} top).
From each \textit{region} we collect four images, one per year from 2018, 2019, 2020 and 2021, covering temporal changes in a given area (\cref{fig:examples_database}, bottom). This produces a database of 700k images.

Unlike our queries, these database images have undergone post-processing that includes atmospheric corrections as well as cloud pixel minimization via composition of images from multiple days. Additionally, the database images are always taken nadir facing (\ie perpendicular to Earth's surface) to minimize obliquity effects . Each database image has resolution 1024$\times$1024.
Further images from the database are shown in \cref{sec:supp_further_examples} of the Supplementary.


\subsection{Evaluation sets}
\label{sec:test_sets}

\input{figures/test_sets_size}

\input{tables/datasets}

\input{figures/batching}
With the end goal of fast and accurate localization of the astronaut photography archive, we propose a deployment environment that takes advantage of the known nadir points and visibility limits associated with each image. So while localization can be performed by matching any given query against the whole world-wide database, a more suitable approach is to follow a divide and conquer paradigm by grouping queries according to their nadir, which allows us to use only a subset of the database for each search, increasing speed and improving results.

We seek to create evaluation sets that mirror the situation seen in deployment, so we evaluate on sets of queries that have a nadir within a 2500 km radius from a chosen Point of Interest (POI).
We then build a database of images such that the database fully encompasses the area that could have been photographed from each collected query, which is an area $2*d_{visible}$ from the original POI, as shown in \cref{fig:test_sets}. 

Since we need the queries' ground truth position for evaluation, we choose queries that have already been localized with exhaustive image matching techniques \cite{Stoken_2023_CVPR}. For these queries, the full photo extent is publicly available\footnote{\url{https://eol.jsc.nasa.gov/SearchPhotos/PhotosDatabaseAPI/} (mlcoord Table)}.

To properly evaluate different methods, we choose six POIs that represent various geographies but all have scientific relevance, and center an evaluation set around each. These are summarized in \cref{tab:datasets}.

%% file: figures/examples_queries.tex
\begin{figure}
    \begin{center}
        \begin{minipage}{.15\textwidth}
            \includegraphics[width=\textwidth]{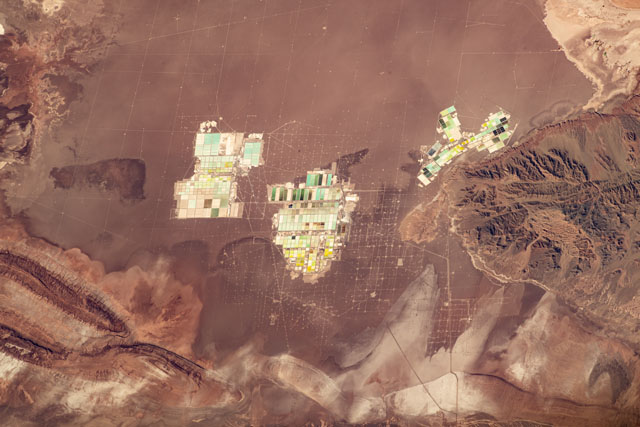}
        \end{minipage}
        \begin{minipage}{.15\textwidth}
            \includegraphics[width=\textwidth]{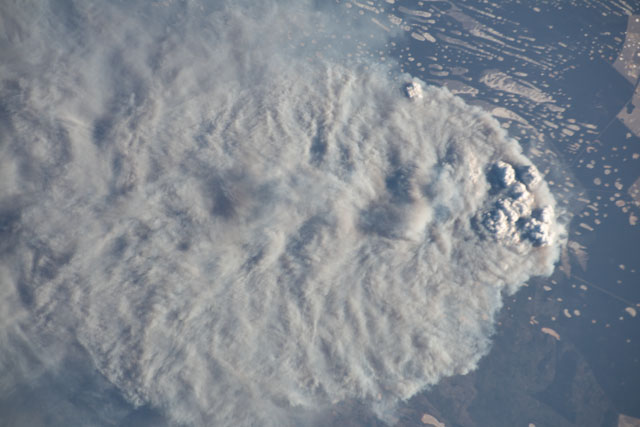}
        \end{minipage}
        \begin{minipage}{.15\textwidth}
            \includegraphics[width=\textwidth]{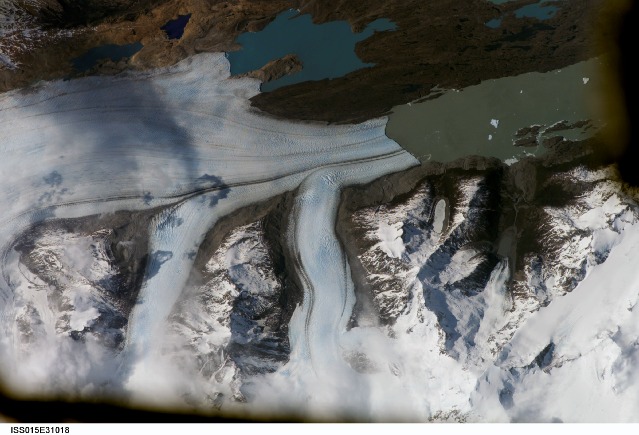}
        \end{minipage}
        \begin{minipage}{.15\textwidth}
            \includegraphics[width=\textwidth]{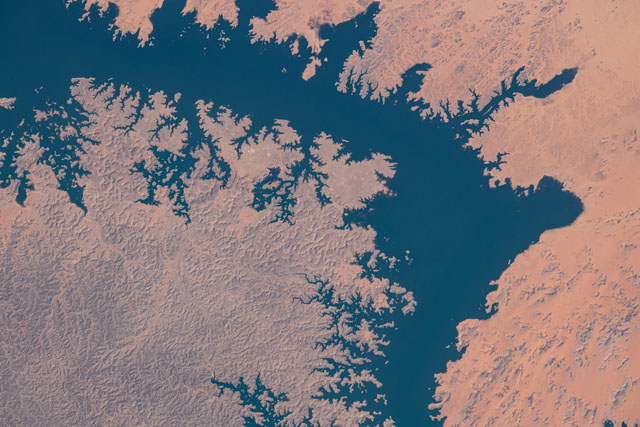}
        \end{minipage}
        \begin{minipage}{.15\textwidth}
            \includegraphics[width=\textwidth]{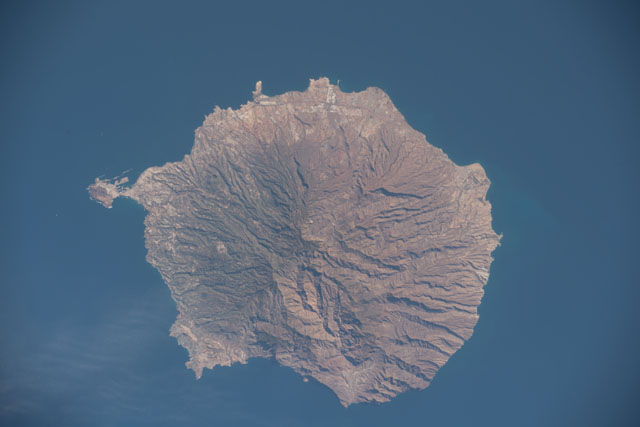}
        \end{minipage}
        \begin{minipage}{.15\textwidth}
            \includegraphics[width=\textwidth]{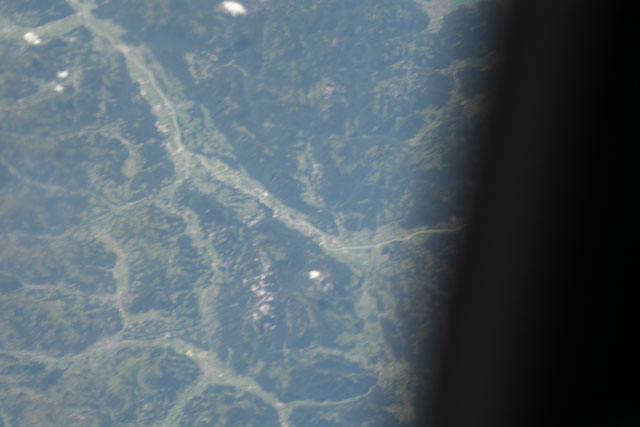}
        \end{minipage}
    \end{center}
\vspace{-3mm}
    \caption{\textbf{Astronaut photo query examples}, showcasing the large variability in covered area and appearance.
    }
\vspace{-3mm}
    \label{fig:examples_queries}
\end{figure}

%% file: figures/examples_database.tex
\begin{figure}
    \begin{center}
        \begin{minipage}[b]{.48\textwidth}
            \includegraphics[width=\textwidth]{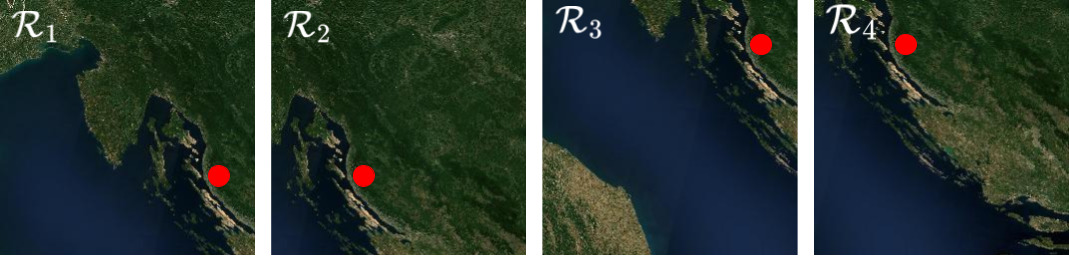}
        \vspace{-2mm}
        \end{minipage}
        \begin{minipage}[t]{.48\textwidth}
            \includegraphics[width=\textwidth]{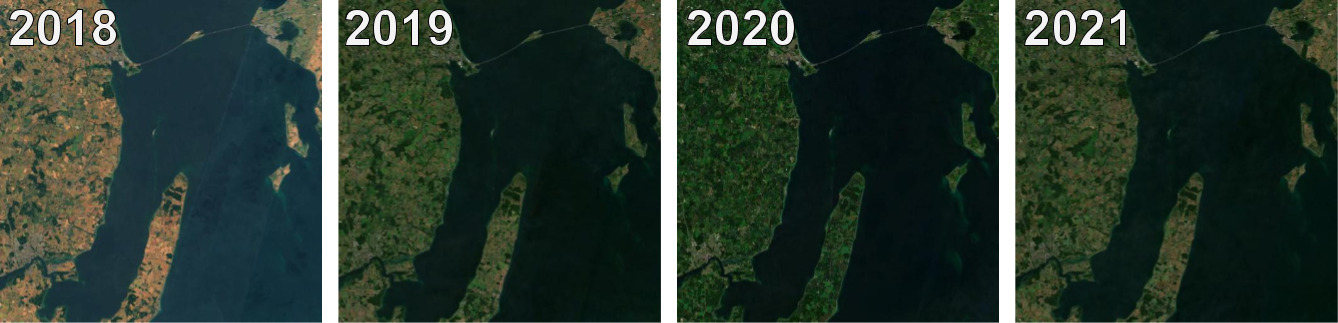}
        \end{minipage}
    \end{center}
    \vspace{-3mm}
    \caption{\textbf{Examples of database images.} \textbf{Top}: four images from different \textit{regions}, with 25\% or 50\% overlap between any pair. The red dot in each image represents the same geographic point. \textbf{Bottom}: 4 images from the same \textit{region} across different years. 
    }
    \vspace{-3mm}
    \label{fig:examples_database}
\end{figure}

%% file: figures/test_sets_size.tex
\begin{figure}
    \begin{center}
    \includegraphics[width=0.999\linewidth]{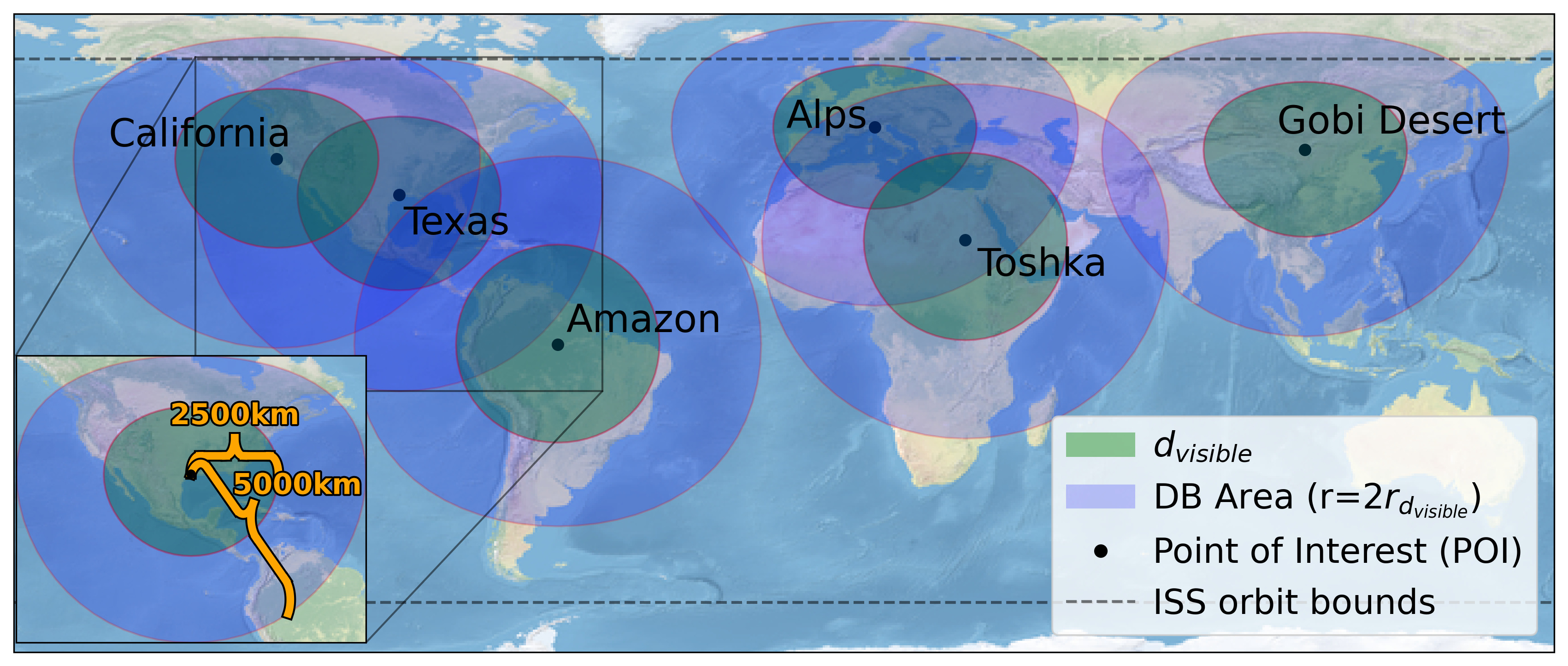}
    \end{center}
    \vspace{-4mm}
    \caption{\textbf{To create evaluation sets} we choose all images that could contain a Point of Interest (POI) - all photos with nadir point within $d_{visible}$. To localize all photos within this range, even if they do not contain the POI, we create a database that contains all areas visible from the nadir points of all selected photos,  yielding a database area of about 5000km$^2$ per POI.
    }
    \label{fig:test_sets}
\end{figure}

%% file: tables/datasets.tex
\begin{table}
\begin{center}
\begin{adjustbox}{width=0.99\columnwidth}
\begin{tabular}{l|cccl}
\toprule
Name & (Lat, Lon) & \# queries & \# DB & Interest \\
\midrule
Texas        &  (30,  -95) & 6142 & 34k & Often photoed (many queries) \\
Alps         &  (45,   10) & 2394 & 53k & Sci. interest -  glacial change \\
California   &  (38, -122) & 3568 & 30k & Disaster response - wildfires \\
Gobi Desert  &  (40,  105) &  726 & 54k & Sci. interest -  desertification \\
Amazon       & (-3,   -60) &  682 & 19k & Sci. interest - deforestation \\
Toshka Lakes &  (23,   30) & 2164 & 63k & Sci. interest -  flood monitoring \\
\bottomrule
\end{tabular}
\end{adjustbox}
\end{center}
\vspace{-5mm}
\caption{\textbf{Evaluation sets.} The six sets cover diverse geographies and relevant areas for research on topics like climate change and disaster management. DB stands for database.}
\vspace{-5mm}
\label{tab:datasets}
\end{table}

%% file: figures/batching.tex
\begin{figure*}
    \begin{center}
    \includegraphics[width=0.88\linewidth]{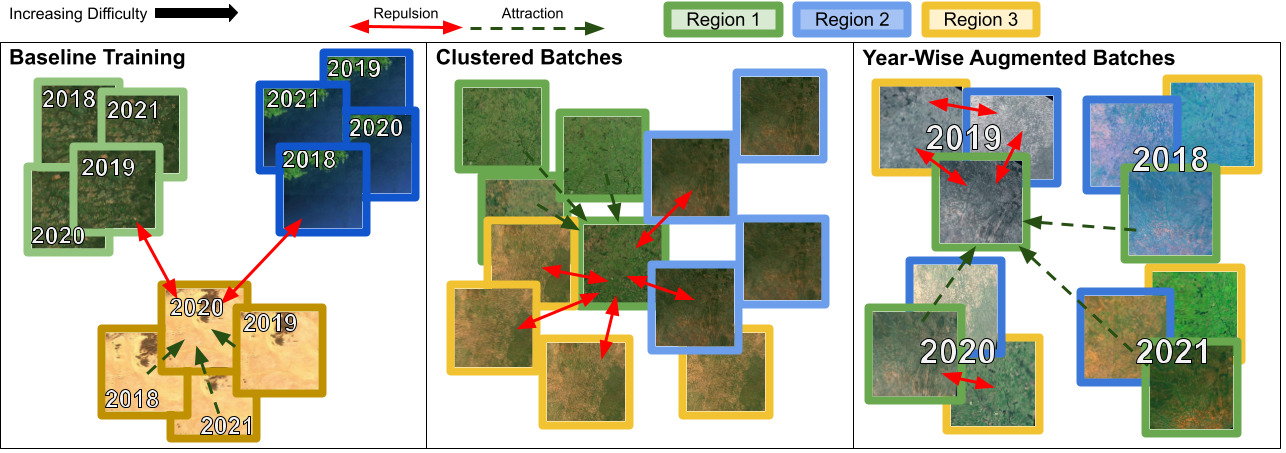}
    \end{center}
    \vspace{-4mm}
    \caption{\textbf{Training strategy.} Naive batching produces too easy a task for the model to learn robust representations from. We increase difficulty by clustering batches by similarity and adding year-wise augmentations, which move images of different \textit{regions}, but from the same year, closer together in feature space. The model must then learn representations that discern similarity in this more difficult context.}
    \vspace{-4mm}
    \label{fig:batching}
\end{figure*}

%% file: sec/4_method.tex
\section{Method}
\label{sec:method}
The goal of this work is to estimate the location of each astronaut photo query, which we tackle through image retrieval: for any given query, the objective is to find the most similar image(s) from the database, and use the corresponding location labels to confidently localize the query.
With this in mind, we aim to train a retrieval model on satellite imagery such that the model is robust to changes in scale, perspective, and color, as well as temporal/seasonal variation.
Specifically, the model's task is to extract features from each image. Then, for each query, the most similar database image is retrieved via kNN in feature space.

In the next subsections we describe how we train our retrieval model using the proposed database.


\subsection{Baseline Training}
\label{sec:baseline_training}
We want our model to extract relevant features of satellite imagery, so a straightforward approach would be to train via contrastive learning: for a given batch of images (\textit{samples}), generate a number of augmented views (typically two), and apply a contrastive loss to maximize the distance in feature space between different samples and minimize distance between different views of the same sample.
However, this solution would not allow the model to learn robustness to temporal changes, because the augmented views come from the same sample, with no inherent temporal variation.

Therefore, instead of producing multiple views of the same image via synthetic augmentation, we train the model by randomly selecting a number of \textit{regions}, and for each \textit{region} we consider its four images (a \textbf{quadruplet}), one per year. Within a batch of quadruplets, the four images from each \textit{region} are mutual positives, whereas any two images from different \textit{regions} are considered negatives: we can therefore consider each \textit{region} as a class. This provides a natural way to address temporal variation.
We can then train the model through a multi-similarity loss \cite{Wang_2019_multi_similarity_loss}.
This idea is shown in \cref{fig:batching} (left).


\subsection{Clustered Batches}
\label{sec:clustered_batches}

The baseline paradigm simply creates batches of random \textit{regions} -- a single batch of quadruplets could contain samples depicting shorelines, islands, deserts, and mountains. The loss operates batchwise, so large intra-batch variability can lead the model to learn too swiftly to distinguish between such diverse \textit{regions}. In these situations, training converges quickly and the model stops short of learning to discriminate between similar-looking, but distinct, \textit{regions}.

To avoid this issue, we want to train with batches of highly similar images, which are still from different \textit{regions}.
To this end, we compute clusters of images with similar representations. Every $N_{iter}$ training iterations, we extract the features from each \textit{region} with the trained model, using only images from a single year, and group them into $C$ clusters.

Then, we create training batches using samples from a given cluster, choosing a different cluster for each batch. This increases the difficulty of the task in each batch, ensuring that the model learns to discriminate between similar-looking images of different \textit{regions}. Note that clusters are not geographically related, but related in feature space. For example, a cluster may contain images from the Gobi and the Sahara deserts, which both appear similar in feature space despite being geographically distant.
This procedure, which can be seen as an offline mining approach,
increases training time by less than 5\% but improves recall by roughly 15\% on average (see \cref{sec:ablations}), a worthwhile trade off.

This concept is illustrated in \cref{fig:batching} (middle), and qualitative examples of images that are clustered together are shown in \cref{sec:supp_clustered_batches} of the Supplementary.


\subsection{Year-Wise Data Augmentation}
\label{sec:year_wise_data_augmentation}

Query images span the ongoing 20+ years of ISS operations, as well as multiple camera types, camera operators, and view points, so queries depicting the same location can have a number of natural ``transformations". These range from blueish hues from lens filters, to skewed representations in highly oblique shots, to seasonal variation in ground cover (e.g. snow, autumn colors).

To make the model robust to such transformations, we employ data augmentation.
Typically, augmentations are applied per image or per batch, but we argue that to truly learn robustness to the augmentations in this setting, the same augmentation should be performed on images from different \textit{regions}, but from the same year.

Formally, at each training iteration 
we choose a set of 4 random augmentations, one for each year $A_{2018}, A_{2019}, A_{2020}, A_{2021}$. We then apply the corresponding augmentation to all images from that year (i.e. $A_{2018}$ is applied to all images from 2018, across all \textit{regions}). Such ``year-wise" augmentation has the effect of moving images from a given year closer in feature space to each other, as shown in \cref{fig:batching} (right). In the loss, images from the same \textit{region}, but different years, are treated as positives, while images from different \textit{regions}
are treated as negatives. This has the effect that images that receive the same augmentation are assigned different classes -- forcing the model to learn to ignore such augmentations -- and instead pull together images from the same \textit{region}, \textit{across} different years.

Furthermore, given that the same augmentation is applied to multiple images, augmentation can be batched and performed on the GPU, making training slightly faster (by 5\% on average, although this depends on the hardware).


\input{tables/main_table}
\subsection{Neutral-Aware Multi-Similarity Loss}
\label{sec:neutral_aware}
\input{figures/dataset_collection_multi_sim}
The powerful Multi-Similarity loss is designed to take as input multiple images from a number of classes, along with the corresponding class indices, and compute a loss from the positives and negatives for each image.

However, it does not take into account that in some cases, pairs of images are neither positives nor negatives, and should be considered ``neutral".
While the concept of neutral images does not exists in standard image retrieval tasks, as an image either belongs to class A or class B, in APL it must be taken into consideration.
For example, should two images, which have an intersection over union of 25\%, be considered positives or negatives? This is precisely the case when two images of the same area, but from different zoom levels, are batched together, as well as overlapping images at the same level (see \cref{fig:dataset_collection_multi_sim}).
We believe the answer is neither, so we add the ``neutral" case to the loss.

Formally, we model the neutral case with the indicator function $\mathcal{I}_{NA}$, which is 0 when the \textit{regions} $\mathcal{R}_i$ and $\mathcal{R}_j$ intersect but are not equal, and 1 otherwise.
\begin{align}
        \mathcal{I}_{NA}(\mathcal{R}_i, \mathcal{R}_j) &= \mathbbm{1}_{(\mathcal{R}_i\cap\mathcal{R}_j)\land (\mathcal{R}_i\neq\mathcal{R}_j)} 
\end{align}
With this, the neutral-aware multi-similarity loss becomes 
\begin{equation}
    \begin{split}
                \mathcal{L}_{NAMS} = \frac{1}{BS}\sum_{i=1}^{BS} \left\{\frac{1}{\alpha} \text{log} \left[ 1+ \sum_{k\in\mathcal{R}_i}e^{-\alpha(\mathcal{S}_{ik}-\lambda)} \right] \right. \\ 
    + \left. \frac{1}{\beta} \text{log} \left[ 1+ \sum_{k\notin\mathcal{R}_i}\mathcal{I}_{NA}(\mathcal{R}_i, \mathcal{R}_k)e^{-\beta(\mathcal{S}_{ik}-\lambda)} \right] \right\}
    \end{split}
\end{equation}
where $BS$ is the number of images in a batch, $\mathcal{R}_i$ is the \textit{region} of image $i$, $\mathcal{S}_{ik}$ is the similarity between image $i$ and $j$, $\lambda$ is a margin, and $\alpha$, $\beta$ are hyperparameters. 

We believe that the neutral-aware multi-similarity loss is applicable to any task where it is not straightforward how to enforce full positivity or negativity between any pair of images, including localization tasks like VPR.

%% file: tables/main_table.tex
\begin{table*}
\begin{center}
\begin{adjustbox}{width=0.999\textwidth}
\begin{tabular}{cc|ccc c ccc c ccc c ccc c ccc c ccc}
\toprule
\multirow{2}{*}{Method} & Type of &
\multicolumn{3}{c}{Texas (validation)} &&
\multicolumn{3}{c}{Alps} &&
\multicolumn{3}{c}{California} &&
\multicolumn{3}{c}{Gobi Desert} &&
\multicolumn{3}{c}{Amazon} &&
\multicolumn{3}{c}{Toshka Lakes} \\
\cline{3-5} \cline{7-9} \cline{11-13} \cline{15-17} \cline{19-21} \cline{23-25}
& Training Imagery
& R@1 & R@10 & R@100 &
& R@1 & R@10 & R@100 &
& R@1 & R@10 & R@100 &
& R@1 & R@10 & R@100 &
& R@1 & R@10 & R@100 &
& R@1 & R@10 & R@100
\\
\midrule
Nadir        & - & 2.4 & - & - && 1.2 & - & - && 2.4 & - & - && 1.8 & - & - && 3.1 & - & - && 1.4 & - & - \\

Random Choice & - &  0.2 &  1.7 & 15.5 &&  0.1 &  1.1 & 11.6 &&  0.2 &  2.3 & 20.1 &&  0.1 &  1.0 & 13.2 &&  0.1 &  1.1 & 11.5 &&  0.2 &  1.2 &  9.1 \\
\midrule

NetVLAD \cite{Arandjelovic_2018_netvlad}   & Urban VPR &  3.6 & 13.4 & 34.8 & &  5.2 & 15.2 & 36.2 & &  4.3 & 14.4 & 39.0 & &  2.1 &  7.7 & 24.5 & &  6.6 & 19.9 & 46.3 & &  9.6 & 22.5 & 45.6 \\
SFRS \cite{Ge_2020_sfrs}                   & Urban VPR &  6.8 & 17.4 & 40.9 & &  7.0 & 17.8 & 41.4 & &  5.4 & 17.2 & 44.4 & &  5.2 & 14.7 & 37.6 & &  9.4 & 23.8 & 52.5 & & 10.8 & 20.4 & 42.4 \\
Conv-AP \cite{Alibey_2022_gsvcities}       & Urban VPR &  3.7 & 13.5 & 37.7 & &  2.0 &  8.8 & 25.0 & &  3.8 & 15.4 & 41.2 & &  3.7 &  9.0 & 24.2 & &  7.0 & 21.1 & 47.7 & &  5.0 & 12.3 & 32.6 \\
CosPlace \cite{Berton_2022_cosPlace}       & Urban VPR &  4.9 & 15.2 & 42.0 & &  5.9 & 19.3 & 45.2 & &  4.8 & 15.2 & 43.3 & &  7.9 & 17.5 & 39.5 & &  8.5 & 23.9 & 54.4 & &  9.8 & 25.4 & 53.0 \\
MixVPR \cite{Alibey_2023_mixvpr}           & Urban VPR &  4.5 & 15.3 & 40.1 & &  3.1 &  8.6 & 22.6 & &  4.4 & 17.5 & 43.2 & &  4.0 & 10.1 & 24.5 & &  9.8 & 28.7 & 58.7 & &  5.9 & 15.4 & 36.5 \\
EigenPlaces \cite{Berton_2023_EigenPlaces} & Urban VPR &  6.4 & 21.6 & 52.4 & &  8.7 & 21.4 & 50.3 & &  6.0 & 22.8 & 54.3 & &  8.1 & 20.0 & 40.9 & & 10.9 & 26.5 & 57.9 & & 14.3 & 30.9 & 60.4 \\
AnyLoc \cite{Keetha_2023_AnyLoc} (DINOv2 \cite{Oquab_2023_dinov2} + NetVLAD) & Universal VPR & \underline{44.1} & \underline{68.7} & \textbf{87.8} && \underline{40.7} & \underline{70.8} & \textbf{92.0} && \underline{48.7} & \textbf{75.0} & \textbf{91.6} && \underline{28.7} & \underline{57.0} & \underline{81.7} && \underline{38.6} & \underline{63.8} & \textbf{86.2} && \underline{63.7} & \textbf{84.5} & \textbf{96.3} \\

\midrule
TorchGeo \cite{Stewart_2022_TorchGeo} (ResNet50 w MOCO \cite{He_2020_moco}) & Satellite &  1.0 &  3.2 & 11.6 &&  0.3 &  1.3 &  5.7 &&  1.1 &  4.3 & 16.4 &&  0.4 &  2.8 &  8.3 &&  1.9 &  6.6 & 19.4 &&  0.7 &  2.9 &  9.9 \\
TorchGeo \cite{Stewart_2022_TorchGeo} (ResNet50 w SeCo \cite{Yao_2021_seco}) & Satellite &  6.1 & 15.6 & 41.7 &&  7.4 & 20.2 & 49.2 &&  5.1 & 14.5 & 37.1 &&  3.7 & 14.0 & 38.9 &&  4.6 & 13.3 & 32.9 &&  5.7 & 15.6 & 38.5 \\
TorchGeo \cite{Stewart_2022_TorchGeo} (ResNet50 w GASSL \cite{ayush2021geography}) & Satellite & \underline{ 9.7} & \underline{22.8} & \underline{46.4} &&  \underline{9.1} & \underline{23.1} & \underline{50.5} && \underline{13.3} & \underline{31.4} & \underline{58.8} &&  \underline{6.3} & \underline{17.5} & \underline{45.4} && \underline{8.3} & \underline{20.3} & \underline{40.1} && \underline{20.4} & \underline{38.6} & \underline{64.2} \\
\midrule
OGCL UAV-View \cite{Deuser_2023_ogcl_uav_view} (ConvNeXt-XXLarge)  & UAV & 16.2 & \underline{35.3} & \underline{65.8} && 14.4 & \underline{34.1} & \underline{64.7} && 19.6 & 42.1 & 71.0 &&  5.5 & 20.5 & 45.3 && 10.0 & 26.3 & 54.7 && 21.1 & 38.4 & 63.4 \\
OGCL UAV-View \cite{Deuser_2023_ogcl_uav_view} (ViT-L/14) \cite{Dosovitskiy_2021_vit} & UAV & \underline{17.6} & 33.2 & 55.9 && \underline{14.6} & 33.2 & 63.9 && \underline{22.8} & \underline{48.1} & \underline{74.9} &&  \underline{7.6} & \underline{22.8} & \underline{50.1} && \underline{20.4} & \underline{39.1} & \underline{62.5} && \underline{31.8} & \underline{51.8} & \underline{74.4} \\
MBEG \cite{Zhu_2023_uav_backbone_winnerUAV_mbeg} (ViT-L/14) & UAV  &  7.0 & 17.6 & 35.1 &&  6.6 & 19.3 & 45.9 &&  8.7 & 20.7 & 41.5 &&  4.4 & 15.0 & 38.0 &&  6.4 & 17.1 & 39.3 &&  8.1 & 20.7 & 49.1 \\
\midrule
EarthLoc (Ours)          & Satellite & \textbf{54.6} & \textbf{72.1} & \underline{87.5} && \textbf{53.9} & \textbf{71.9} & \underline{87.2} && \textbf{55.9} & \underline{74.6} & \textbf{91.6} && \textbf{46.8} & \textbf{65.0} & \textbf{82.9} && \textbf{45.6} & \textbf{66.6} & \underline{82.4} && \textbf{67.6} & \underline{80.3} & \underline{91.9} \\
\bottomrule
\end{tabular}
\end{adjustbox}
\end{center}
\vspace{-5mm}
\caption{\textbf{Results from different methods on our query sets.} Methods are grouped into (1) naive, (2) VPR, (3) Remote Sensing (UAV+Satellite), (4) other baselines trained on our database with clustered batches (\cref{sec:clustered_batches}). For all methods we use \textit{4x90TTA}. VPR stands for Visual Place Recognition, UAV for Unmanned Aerial Vehicle. Best results overall in \textbf{bold}, best per-group \underline{underlined}.
EarthLoc outperforms other models, and performs competitively with AnyLoc, while having 50 times faster features extraction and 10 times smaller features.
}
\vspace{-5mm}
\label{tab:main_table}
\end{table*}

%% file: figures/dataset_collection_multi_sim.tex
\begin{figure}
    \begin{center}
    \includegraphics[width=0.7\columnwidth]{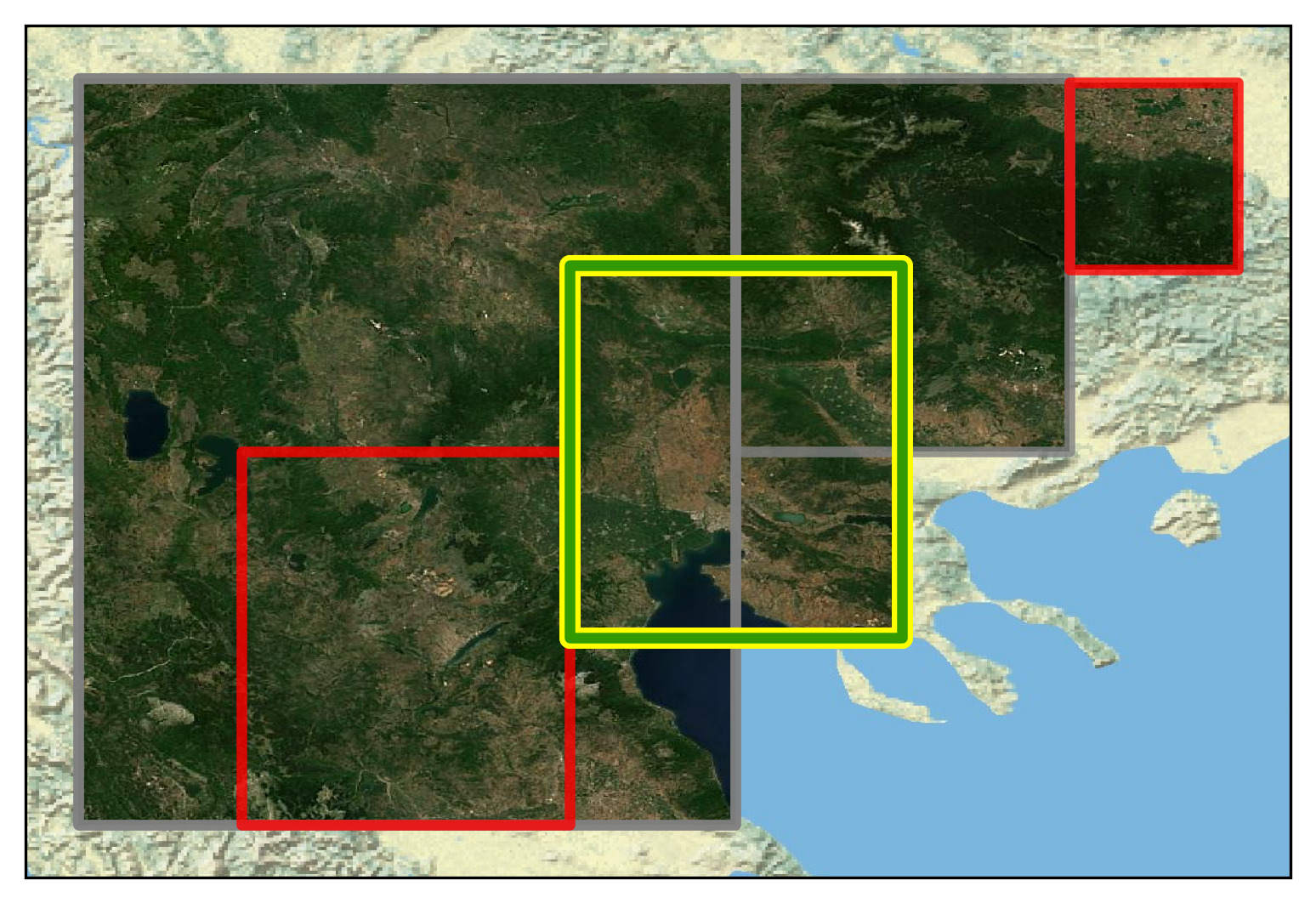}
    \end{center}
\vspace{-5mm}
    \caption{\textbf{Dataset collection from satellite imagery and handing in  Neutral-Aware Multi-Similarity Loss.} For a given image (yellow), images with no overlap are negatives (red), images from the same \textit{region} are positives (green, overlapping yellow), and images with partial overlap are neutral (grey).
    Note that the database is made of images at 3 different zooms (each double in size than the next), which explains the \textit{regions} of different size.
    }
\vspace{-6mm}
    \label{fig:dataset_collection_multi_sim}
\end{figure}

%% file: sec/5_experiments.tex
\section{Experiments}
\label{sec:experiments}

\subsection{Experimental setting}
\subsubsection{Training}
In training, we use a batch size of 32 quadruplets (128 images), where one quadruplet corresponds to four images within a \textit{region}. 
Our model has a MixVPR-style architecture \cite{Alibey_2023_mixvpr}, with a ResNet50 \cite{He_2016_resnet} backbone and an MLP-Mixer\cite{Tolstikhin_2021_mlp_mixer} with output dimension 4096.
We train for 50k iterations with the Adam optimizer \cite{Kingma_2014_adam} and learning rate 0.0001.  Clustering (see \cref{sec:clustered_batches}) is performed every 5k iterations with the number of clusters $C=200$.
As augmentation, we use torchvision's color jittering, random perspective, and random rotation.
Training takes between 10 and 12 hours on an A100 GPU for all methods described in \cref{sec:method}.

\subsubsection{Inference}
Due to the imagery acquisition conditions, we cannot assume any canonical orientation for the photos (in space, there's no gravity vector, \ie, no ``up" or ``down").

This poses a challenge for any retrieval method -- to address it, we generate four feature embeddings for each database image, one each after rotating the image for angles $r\in{[0,90,180,270]}$, in a process we call 4 x 90° test time augmentation (\textit{4x90TTA}).
For fairness, we do this for all methods in all experiments, given that the models are not trained to be robust to such large arbitrary rotations.
This increases latency by 4x, but the method is still orders of magnitude faster than previous approaches (retrieval takes 0.05 seconds per query with \textit{4x90TTA}, while previous works required $\sim$1 minute \cite{Stoken_2023_CVPR} per query). Memory requirements also increase by four times, although this is far from being an issue for a standard workstation, since all database features for the whole Earth fit in less than 12GB of RAM.
Experiments without \textit{4x90TTA} are shown in the Supplementary in \cref{sec:supp_empirical_failure_cases}.
As an added bonus, since each feature from the database refers to an image with a certain orientation (w.r.t. North), retrieval predictions also give an estimate of the orientation of the query.
We use features from the 2021 satellite imagery as our global database.

\paragraph{Success Criteria} With the final goal of localizing the 4.5M astronaut photograph queries in mind, we define a predicted image as correct if there is non-zero overlap between the predication and the query.
As a metric we use Recall@N, defined as the percentage of queries for which at least one of the top-N predictions is correct.


\input{tables/ablation}

\subsection{Results}
In \cref{tab:main_table} we present results from a large number of models from multiple domains.
First, we show the recall when naively predicting the query to be directly nadir, proving that the vast majority of images are taken at an angle. Second, we show the poor performance of randomly choosing a prediction from the database - this is a symptom of the large-scale nature of the task (\ie, localizing a 5000 sq.km wide image within a 20 million sq.km area).

Results with methods trained for Urban Visual Place Recognition (VPR) provide, somewhat unsurprisingly, poor results - on the other hand, AnyLoc \cite{Keetha_2023_AnyLoc} shows remarkably strong results, correctly predicting the location of the query as the first prediction for roughly 50\% of the queries across all evaluation sets, although at the cost of over 50x slower feature extraction and 10x bigger features than its VPR counterparts (AnyLoc uses 49152-D features vs 4096 of NetVLAD).

We then test a number of models trained with self-supervised techniques, like MOCO \cite{He_2020_moco}, SeCo \cite{Yao_2021_seco}, and Geography-aware self-supervised learning (GASSL) \cite{ayush2021geography}, conveniently provided by the TorchGeo library \cite{Stewart_2022_TorchGeo}. Surprisingly, these methods did not provide competitive results, and are on average about equal with Urban VPR methods.

From the Unmanned Aerial Vehicle (UAV) domain, we use the latest state of the art (SOTA) methods, \ie, the winners of the recent challenge in UAV localization at ACM MM (October 2023) \cite{Zheng_2023_UAV_workshop}, namely OGCL \cite{Deuser_2023_ogcl_uav_view} and MBEG \cite{Zhu_2023_uav_backbone_winnerUAV_mbeg}.
Despite being SOTA in the related task of UAV localization, we see that these methods are not competitive with AnyLoc, despite outperforming most other methods.

Finally, EarthLoc outperforms all previous works, localizing 57.1\% of queries within the top-1 prediction (Recall@1) averaged on the 6 evaluation sets, while having 50 times faster extraction time than the second best method (AnyLoc) and ten times smaller features.


\subsection{Ablations}
\label{sec:ablations}
In \cref{tab:ablation} we ablate the components of our method to better understand performance.
Results show that not only do Clustered Batches, Year-Wise Data Augmentation and Neutral-Aware Multi-Similarity Loss provide significant increases over the baseline, but they are also orthogonal improvements and can be easily combined.


\subsection{Qualitative Analysis and Failure Cases}
A sample of qualitative results from EarthLoc are shown in \cref{fig:qualitatives}.
The samples show that EarthLoc picks locations that have similar characteristics to the query (top left), is robust to rotations and changes in scale (bottom right), and provides correct predictions that are difficult to correctly match even for the human eye (top left, bottom left).
While a number of failure cases arise, we could not identify a repetitive pattern for wrongly localized queries, and believe that many of these failure cases can be addressed by better architectures and training paradigms.
More qualitative results are shown in \cref{sec:supp_qualitative_results} of the Supplementary.

\input{figures/qualitatives}


\subsection{Limitations and Future Work}
While this work takes a major step forward in the study of APL, EarthLoc focuses on daytime queries, and we did not investigate panoramic or nighttime imagery. The localization of nighttime astronaut photos in particular is important (e.g. to study light pollution \cite{DeMiguel_2014_night_atlas}), but 
we believe this to be a separate challenge that should be addressed in future work.
Another limitation is due to the zoom limits of the database: we chose database tiles at zoom levels 9, 10, 11 (see \cref{sec:database}), which allows us to localize queries of area between 5000 and 900,000 square km. Queries with smaller or bigger covered areas would need finer or coarser database images to be geolocated.

The originality of this work opens up a large number of possible future research directions, like training directly on (some of the) localized queries, using seasonal data, post-processing predictions, obtaining pixel-wise geolocation, applying domain adaptation, using available metadata (like camera lens) during training, et cetera.

%% file: tables/ablation.tex
\begin{table*}
\begin{center}
\begin{adjustbox}{width=0.95\textwidth}
\begin{tabular}{ccccc | c c cc c cc c cc c cc cccccccccccccc cccc}
\toprule
Multi-similarity &Using \textit{regional}&Clustered& Year-Wise        & Neutral-Aware & 
\multicolumn{2}{c}{Texas (validation)} & &
\multicolumn{2}{c}{Alps} & &
\multicolumn{2}{c}{California} & &
\multicolumn{2}{c}{Gobi Desert} & &
\multicolumn{2}{c}{Amazon} & &
\multicolumn{2}{c}{Toshka Lakes} & &
\multicolumn{2}{c}{Average}
\\
\cline{6-7} \cline{9-10} \cline{12-13} \cline{15-16} \cline{18-19} \cline{21-22} \cline{24-25}   
Loss             & quadruplets           & Batches &Data Augmentation&Multi-Similarity
& R@1 & R@100 &
& R@1 & R@100 &
& R@1 & R@100 &
& R@1 & R@100 &
& R@1 & R@100 &
& R@1 & R@100 &
& R@1 & R@100 \\
\midrule
\checkmark &            &            &            &            & 25.5 & 71.4 & & 24.8 & 77.3 & & 27.8 & 76.6 & & 16.8 & 68.8 & & 23.6 & 61.5 & & 38.8 & 80.8 & & 26.2 & 72.7 \\
\checkmark & \checkmark &            &            &            & 45.0 & 84.1 & & 45.4 & 87.3 & & 49.5 & 88.5 & & 35.7 & 79.9 & & 34.2 & 80.2 & & 62.9 & 91.1 & & 45.4 & 85.2 \\
\checkmark & \checkmark & \checkmark &            &            & 51.4 & 87.5 & & 49.6 & 89.5 & & 53.9 & 90.6 & & 40.8 & 84.3 & & 39.3 & 82.8 & & 64.0 & 93.0 & & 49.8 & 88.0 \\
\checkmark & \checkmark & \checkmark & \checkmark &            & 54.6 & 87.5 & & 53.9 & 87.2 & & 55.9 & \textbf{91.6} & & 46.8 & 82.9 & & 45.6 & 82.4 & & 67.6 & 91.9 & & 54.1 & 87.2 \\
           & \checkmark & \checkmark &            & \checkmark & 55.3 & 88.2 & & 55.1 & \textbf{90.0} & & 57.2 & 90.3 & & 44.6 & 85.8 & & \textbf{48.4} & \textbf{84.8} & & 68.9 & 92.9 & & 54.9 & 88.7 \\
           & \checkmark & \checkmark & \checkmark & \checkmark & \textbf{55.9} & \textbf{88.3} & & \textbf{58.4} & 89.5 & & \textbf{58.0} & 91.4 & & \textbf{51.1} & \textbf{86.5} & & 47.2 & 84.6 & & \textbf{72.2} & \textbf{93.3} & & \textbf{57.1} & \textbf{88.9} \\

\bottomrule
\end{tabular}
\end{adjustbox}
\end{center}
\vspace{-4mm}
\caption{\textbf{Ablations over test sets.} EarthLoc uses Clustered Batches (\cref{sec:clustered_batches}), Year-Wise Data Augmentation (\cref{sec:year_wise_data_augmentation}) and Neutral-Aware Multi-Similarity Loss (\cref{sec:neutral_aware}). Best results in \textbf{bold}.
}
\vspace{-3mm}
\label{tab:ablation}
\end{table*}

%% file: figures/qualitatives.tex
\begin{figure}
    \centering

    \begin{subfigure}[b]{0.22\textwidth}
        \includegraphics[width=\textwidth]{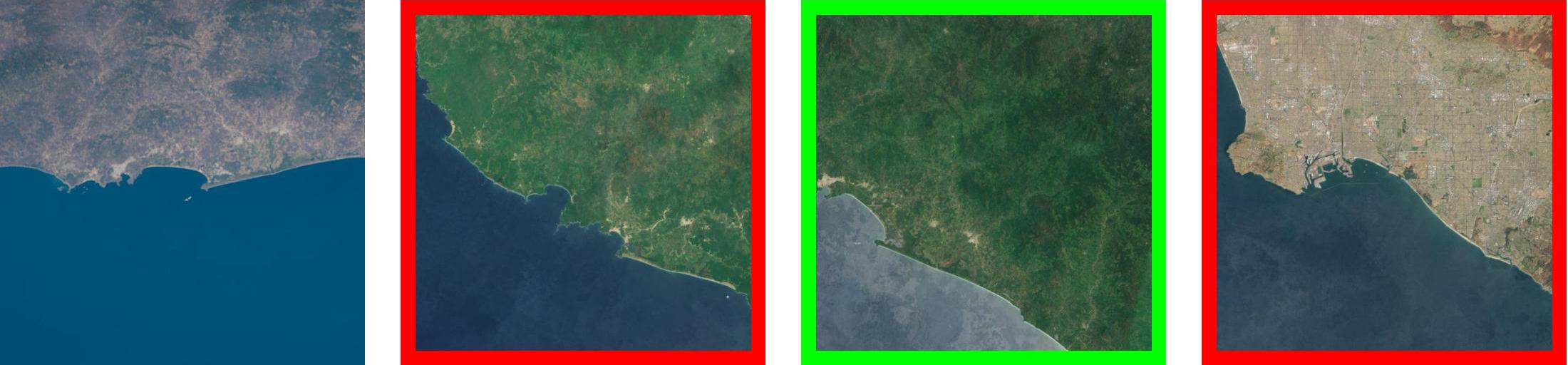}
    \end{subfigure}
    \hfill
    \begin{subfigure}[b]{0.22\textwidth}
        \includegraphics[width=\textwidth]{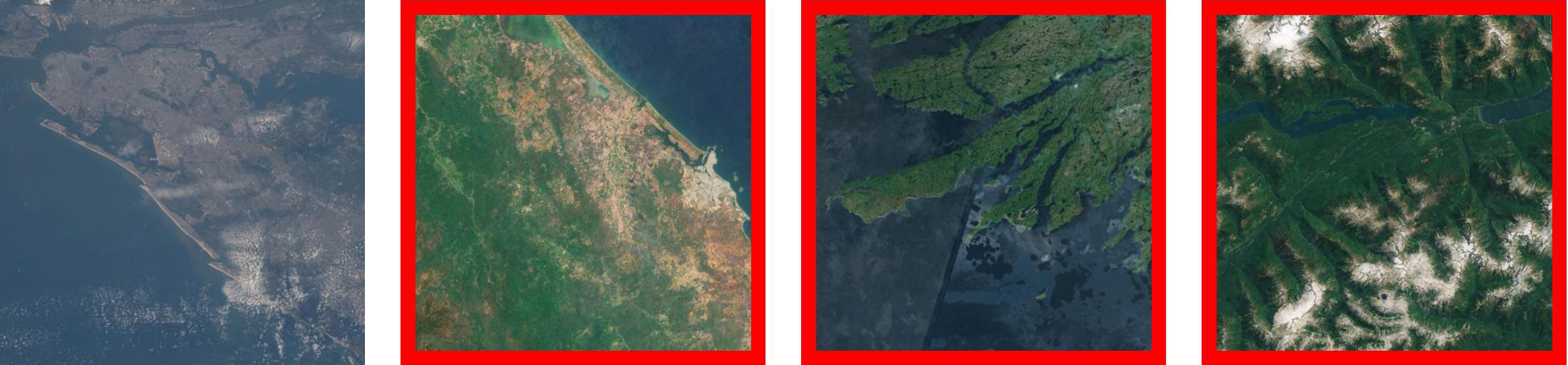}
    \end{subfigure}

    \vspace{8pt} 
    \begin{subfigure}[b]{0.22\textwidth}
        \includegraphics[width=\textwidth]{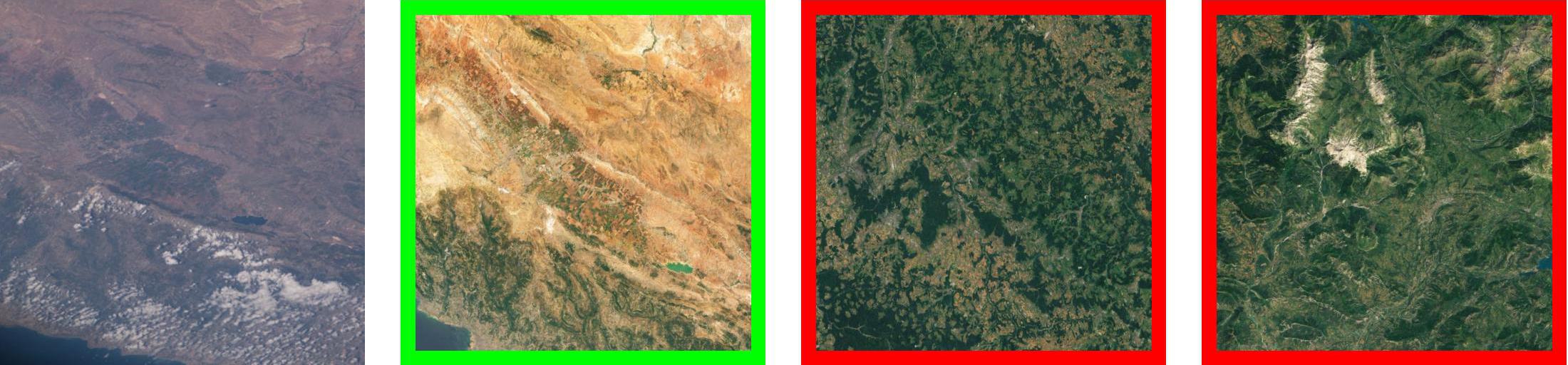}
    \end{subfigure}
    \hfill
    \begin{subfigure}[b]{0.22\textwidth}
        \includegraphics[width=\textwidth]{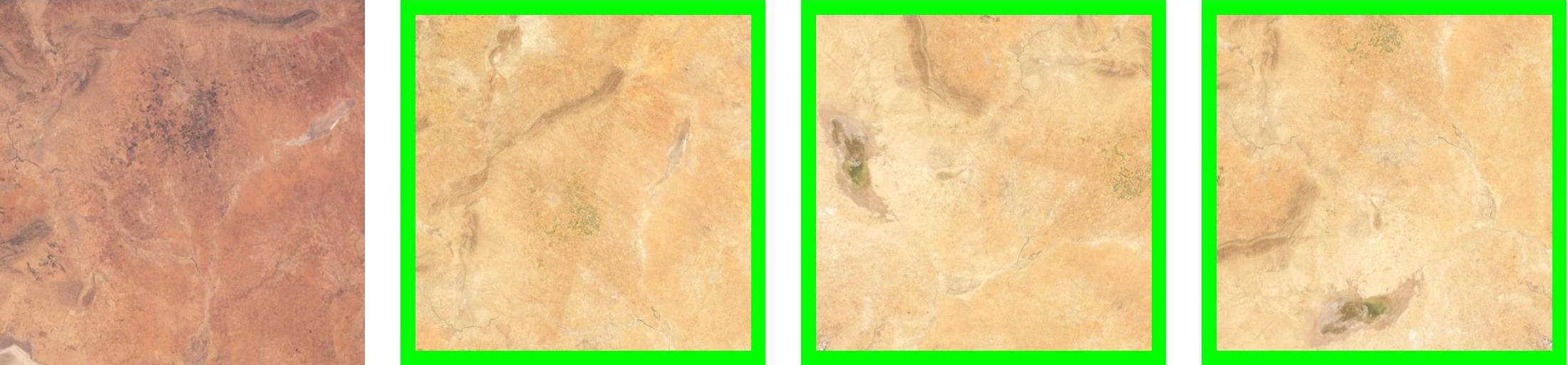}
    \end{subfigure}
    \caption{\textbf{Qualitative results.} Each group of four images represents a query (left) and its top-3 predictions. Green and red indicate if the prediction is correct or incorrect.}
    \vspace{-3mm}
    \label{fig:qualitatives}
\end{figure}

%% file: sec/6_conclusions.tex
\section{Conclusions}
\label{sec:conclusions}
We introduce EarthLoc, a novel method for localizing astronaut photography of Earth. By reframing APL from an image matching task to an image retrieval task,  and introducing a novel loss function and training scheme, our method accurately and efficiently determines the geographic location of astronaut photography given only the position of the International Space Station at the time the photo was taken. Our main contributions are the Neutral-Aware Multi-Similarity Loss, Year-Wise Data Augmentation technique, and new astronaut photography-oriented validation and test datasets to encourage future work on this problem.
\\

\noindent\textbf{Acknowledgements.}
\small{We acknowledge the CINECA award under the ISCRA initiative, for the availability of high performance computing resources.
This work was supported by CINI.
Project supported by ESA Network of Resources Initiative.
This study was carried out within the project FAIR - Future Artificial Intelligence Research - and received funding from the European Union Next-GenerationEU (PIANO NAZIONALE DI RIPRESA E RESILIENZA (PNRR) – MISSIONE 4 COMPONENTE 2, INVESTIMENTO 1.3 – D.D. 1555 11/10/2022, PE00000013). This manuscript reflects only the authors’ views and opinions, neither the European Union nor the European Commission can be considered responsible for them.
European Lighthouse on Secure and Safe AI – ELSA, HORIZON EU Grant ID: 101070617
}

%% file: sec/X_suppl.tex
\clearpage
\setcounter{page}{1}
\maketitlesupplementary

\section{Empirical Investigation of Challenge Modes}
\label{sec:supp_empirical_failure_cases}
The astronaut photography localization task has various challenge modes, where particular photography conditions make localization more difficult for one image compared to another. In this section, we analyze the correlation between two of these conditions (challenge modes) and performance.
\paragraph{Distance from Nadir} As the distance between the ISS nadir point and photo location increases, so too does obliquity and shear due to the imaging geometry. Oblique imagery is often taken through a thicker column of atmosphere, adding blurriness to the image. Thus, we expect some performance drop to accompany increasing distance, as seen beyond 400km in \cref{fig:supp_corr_recall_dist_nadir}. Further augmentation to satellite images during training can potentially close this gap, particularly if these augmentations are designed to simulate conditions seen in far-from-nadir imagery.
\paragraph{Area} We next analyze the correlation between an astronaut photo's geographic area encompassed (area) and recall (\cref{fig:supp_corr_recall_area}). Here, recall decreases as area increases. Based on our training set construction (see \cref{sec:database}), there are fewer images with larger areas required to cover the extent (land area between $\pm60$° latitude), so fewer such images are included in training. We expect including more such images during training will improve performance for this type of imagery.  
\section{Effect of 4x90 TTA}
\label{sec:supp_4x90TTA}
We also study the impact of our \textit{4x90TTA} strategy, with results reported in \cref{tab:4x90TTA}.  In each evaluation set, performance improves due to \textit{4x90TTA}. Based on orientation information from the Gateway to Astronaut Photography of Earth, there are approximate the same number of images across orientation angles in [0°,360°). However, the performance does not increase in proportion to the TTA (i.e., by 4x), indicating that EarthLoc learns some rotation invariance during training. This observation is further supported by multiple rotations of the same image present in the top 3 predictions, examples of which can be seen in \cref{fig:supp_qualitatives}. Note that even in such cases, the database image with the closest orientation is usually the top prediction, followed by correct database images with other orientations.

Despite EarthLoc's ability to retrieve correct images with other orientations, the ablation experiment in \cref{tab:4x90TTA} shows that this capability is limited, and that \textit{4x90TTA} significantly boosts recall on all evaluation sets.
\input{tables/supp_4x90TTA}

\input{figures/supp_corr_recall_dist_nadir}

\input{figures/supp_corr_recall_area}

\section{Features visualizations}
\begin{figure}
    \begin{center}
    \includegraphics[width=0.99\linewidth]{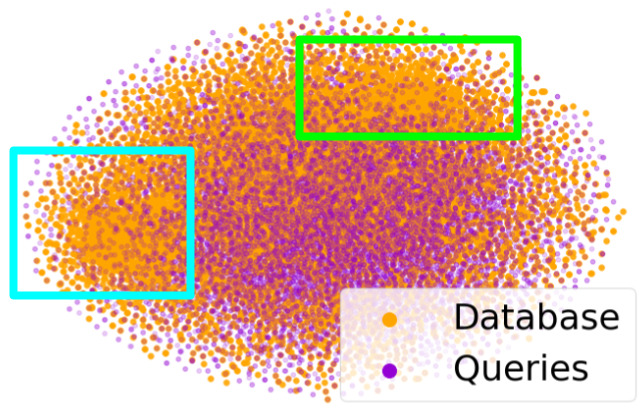}
    \end{center}
    \caption{\textbf{T-SNE representation.} The two colored boxes represent geography types with few queries (areas rarely photographed by astronauts), namely open sea (light blue box) and plains (green).}
    \label{fig:tsne}
\end{figure}

Given the domain gap existing between database and queries images, we show in \cref{fig:tsne}
the distribution of their features, as extracted by EarthLoc, through a T-SNE.

\section{Further Examples of Queries and Database}
\label{sec:supp_further_examples}
\input{figures/supp_queries_db}
We present additional examples of astronaut photo queries (\cref{fig:supp_queries}) and satellite database images (\cref{fig:supp_db}).

These example queries illustrate the variety inherent in astronaut photography. Some images have occlusion due to spacecraft hardware, some images contain clouds, and all have different orientations with respect to North. There is significant geographic and scale variation, with some photos highlighting neighborhoods of large cities and others showing entire lakes or vast mountain ranges. 

The satellite imagery is more regular. In addition to uniform orientation, these images are algorithmically post-processed and thus share similar characteristics, which are somewhat different from those of astronaut photography. Though not all satellite imagery is cloudfree, this particular set is constructed to minimize clouds, and consequently few clouds are seen in the example images. 

\section{Qualitative Results}
\label{sec:supp_qualitative_results}

\subsection{Examples of Queries and Predictions}
\label{sec:supp_qualitatives}
\input{figures/supp_qualitatives}
Example queries and associated top 3 predictions from EarthLoc are in \cref{fig:supp_qualitatives}. Often, multiple orientations of the same database image are within the top 3. In other cases, correct database images with different areas (scale) are retrieved. 

In examples where no correct database image is retrieved, similar looking images are often returned (left, third from bottom). In some failure cases, however, predictions do not show significant similarity to the query (bottom left). We have not found any unifying characteristics in such scenarios.

\subsection{Examples of Clustered Batches}
\label{sec:supp_clustered_batches}
\input{figures/supp_clusters}
As described in \cref{sec:clustered_batches}, we construct batches from clusters to facilitate training. Samples of such clusters are in \cref{fig:supp_clusters}. Clusters are built by collecting \textit{regions} that have similar representations in feature space, despite being from potentially far reaching places on Earth. These similar representations often correspond to shared characteristics, and we can assign high level labels to clusters, like ``rivers in the forest" to the top cluster and ``mountainous deserts" for the second. 

During training, the loss function works to separate the representations for different classes/\textit{regions} that are batched together, so building batches from clusters provides a much more challenging optimization task than random batching, as intra-cluster (across the row) images are much more similar than inter-cluster images (down the columns). 

\subsection{Examples of Year-Wise Data Augmentation}
\label{sec:supp_year_wise_da}
\input{figures/supp_yw_augm}
We show a subset of a training batch as it is presented to the model (i.e., after augmentation) in \cref{fig:supp_yw_augm}. This illustrates our Year-Wise Data Augmentation. Each half-row (4 images) is a \textit{region} quadruplet, with one image from each of the years 2018, 2019, 2020, and 2021. Columns are arranged by year (i.e., each column contains images from the same year). According to our Year-Wise Data Augmentation scheme (see \cref{sec:year_wise_data_augmentation}), images from the same year (i.e., in each column) have received the same augmentation. Augmentations are color jittering, random perspective, and random rotation.

%% file: tables/supp_4x90TTA.tex
\begin{table}
\begin{center}
\begin{adjustbox}{width=0.99\columnwidth}
\begin{tabular}{l|cccccc}
\toprule
& \multicolumn{6}{c}{Recall@1} \\
\cline{2-7} 
Test Time Aug.   & Texas       & Alps        & California  & Gobi        & Amazon      & Toshka      \\
\midrule
None             &  32.6       & 30.9        & 31.1        & 30.4        & 27.2        & 39.1        \\
\textit{4x90TTA} &\textbf{54.6}&\textbf{53.9}&\textbf{55.9}&\textbf{46.8}&\textbf{45.6}&\textbf{67.6}\\

\bottomrule
\end{tabular}
\end{adjustbox}
\end{center}
\vspace{-3mm}
\caption{\textbf{Ablation on \textit{4x90TTA}.} Performance of EarthLoc on each evaluation set with and without \textit{4x90TTA}. \textit{4x90TTA} approximately doubles Recall@1 across the sets.}
\vspace{-3mm}
\label{tab:4x90TTA}
\end{table}

%% file: figures/supp_corr_recall_dist_nadir.tex
\begin{figure}
    \begin{center}
    \includegraphics[width=0.9\linewidth]{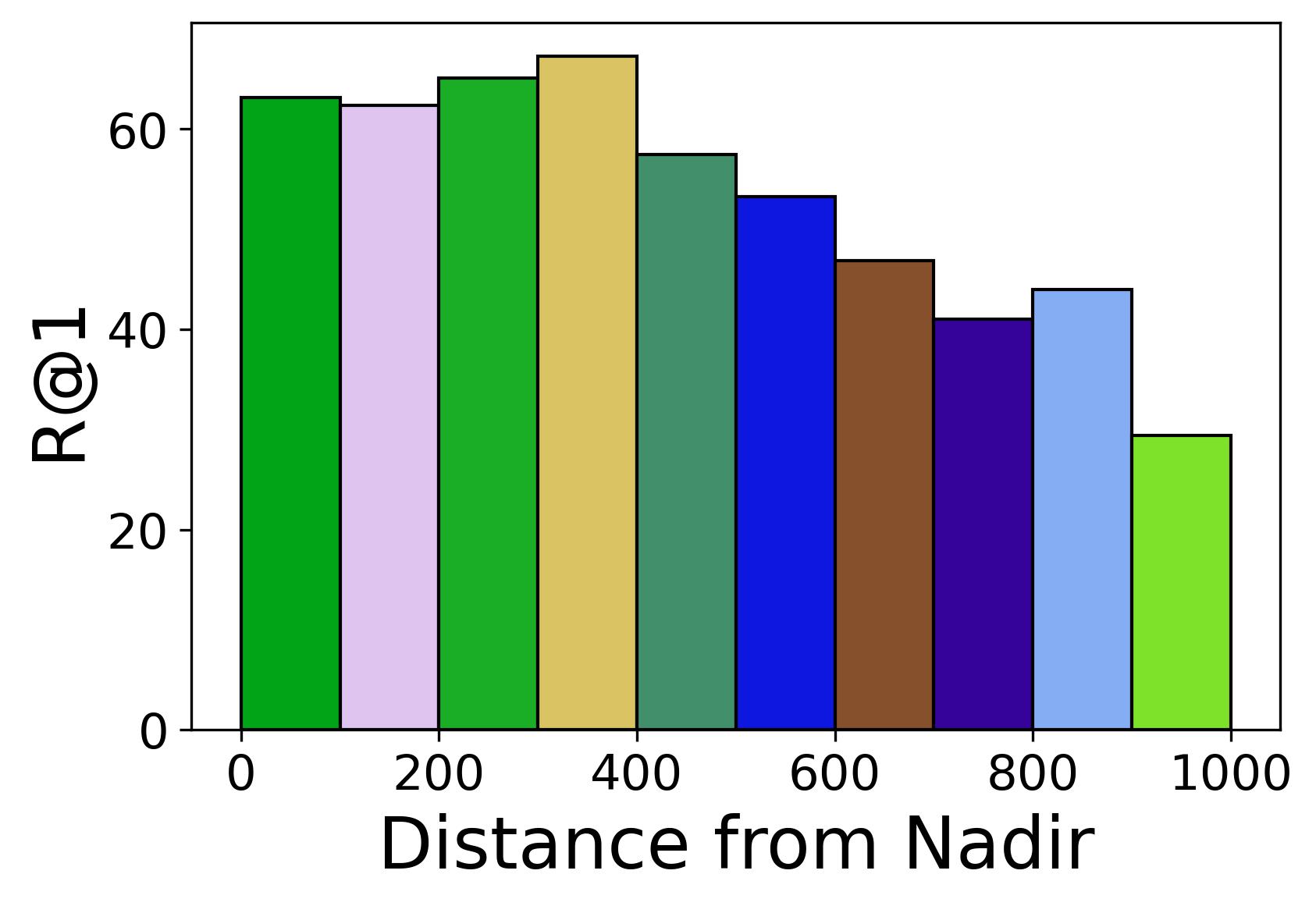}
    \caption{\textbf{Recall@1 vs Distance from Nadir (km). }Performance of EarthLoc as distance from nadir increases. Larger distances are more challenging due to obliquity effects, and this is reflected in the drop in performance.}
    \label{fig:supp_corr_recall_dist_nadir}
    \end{center}
\end{figure}

%% file: figures/supp_corr_recall_area.tex
\begin{figure}
    \begin{center}
    \includegraphics[width=0.9\linewidth]{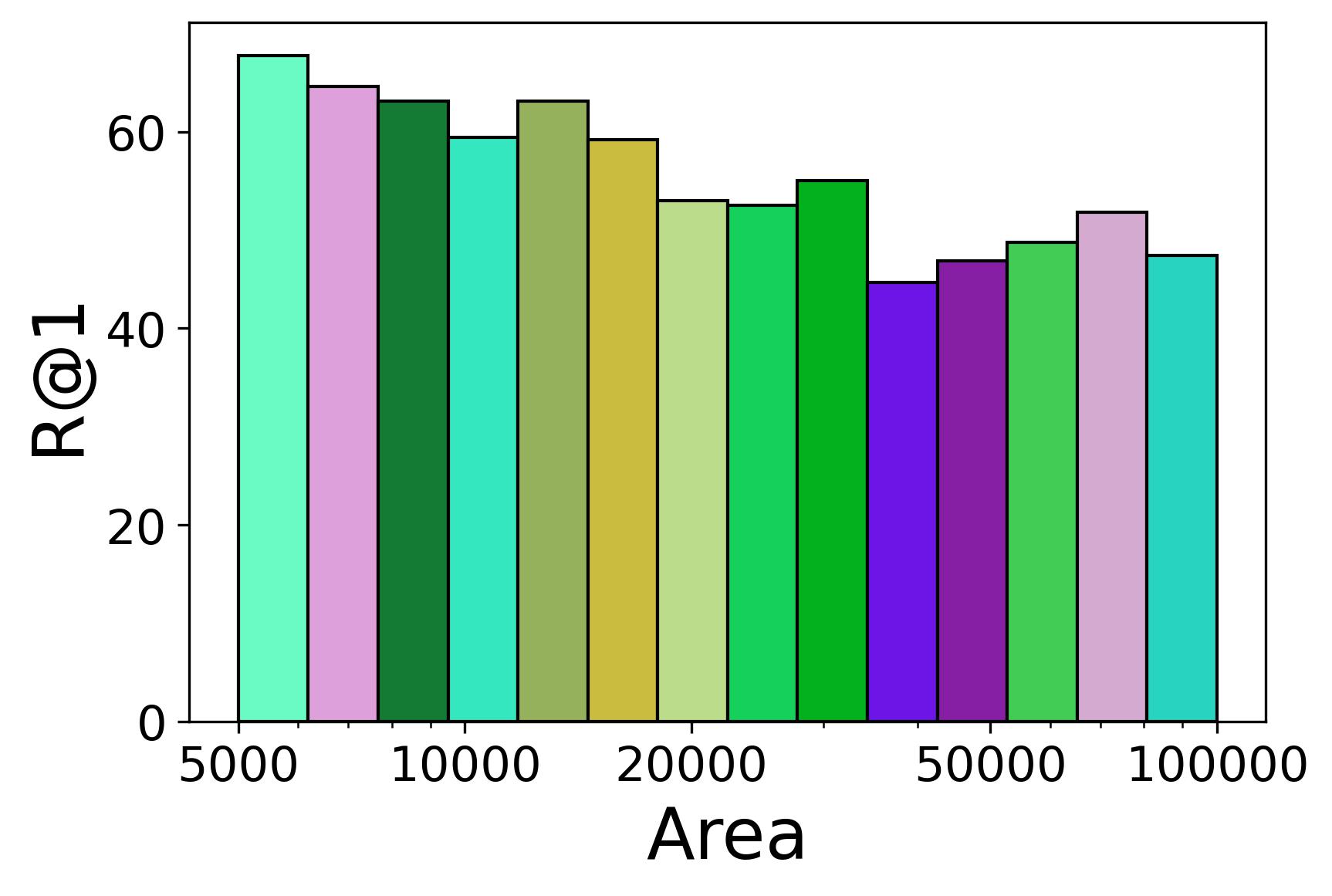}
    \caption{\textbf{Recall@1 vs Area (sq. km).} Performance of EarthLoc on astronaut photo queries with different areas. Decrease in performance for larger area photos is attributed to lower quantities of training data for this regime.}
    \label{fig:supp_corr_recall_area}
    \end{center}
\end{figure}

%% file: figures/supp_queries_db.tex
\begin{figure*}
    \begin{center}
    \includegraphics[width=0.85\textwidth]{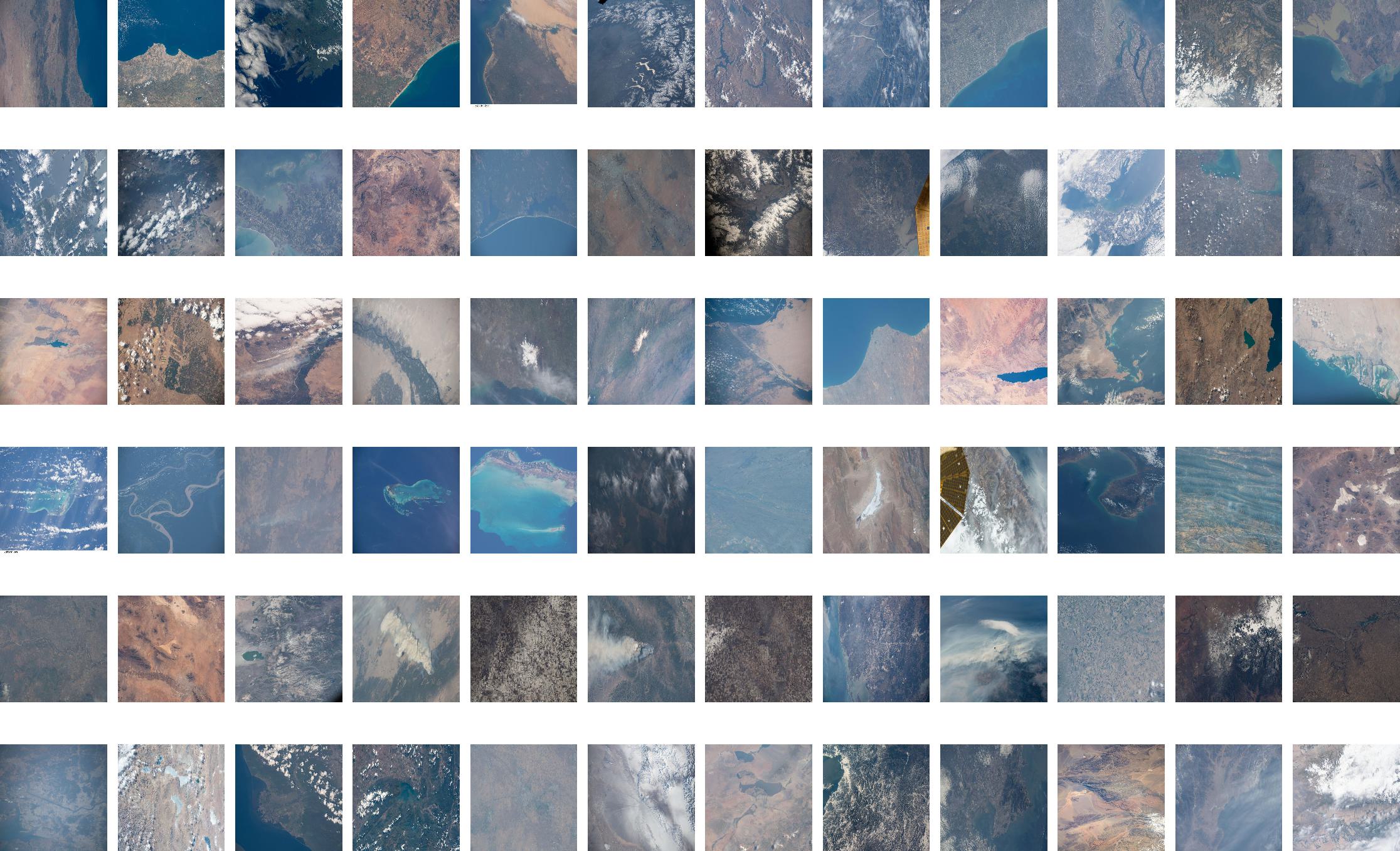}
    \caption{\textbf{Random examples of queries.}
    All queries come from the Gateway to Astronaut Photography of Earth collection \protect\footnotemark.
    Each row is a randomly selected set of queries from each test set, respectively being Alps, Texas, Toshka Lakes, Amazon, California, Gobi.
    }
    \label{fig:supp_queries}
    \end{center}
\end{figure*}
\footnotetext{\url{https://eol.jsc.nasa.gov}}

\begin{figure*}
    \begin{center}
    \includegraphics[width=0.8\textwidth]{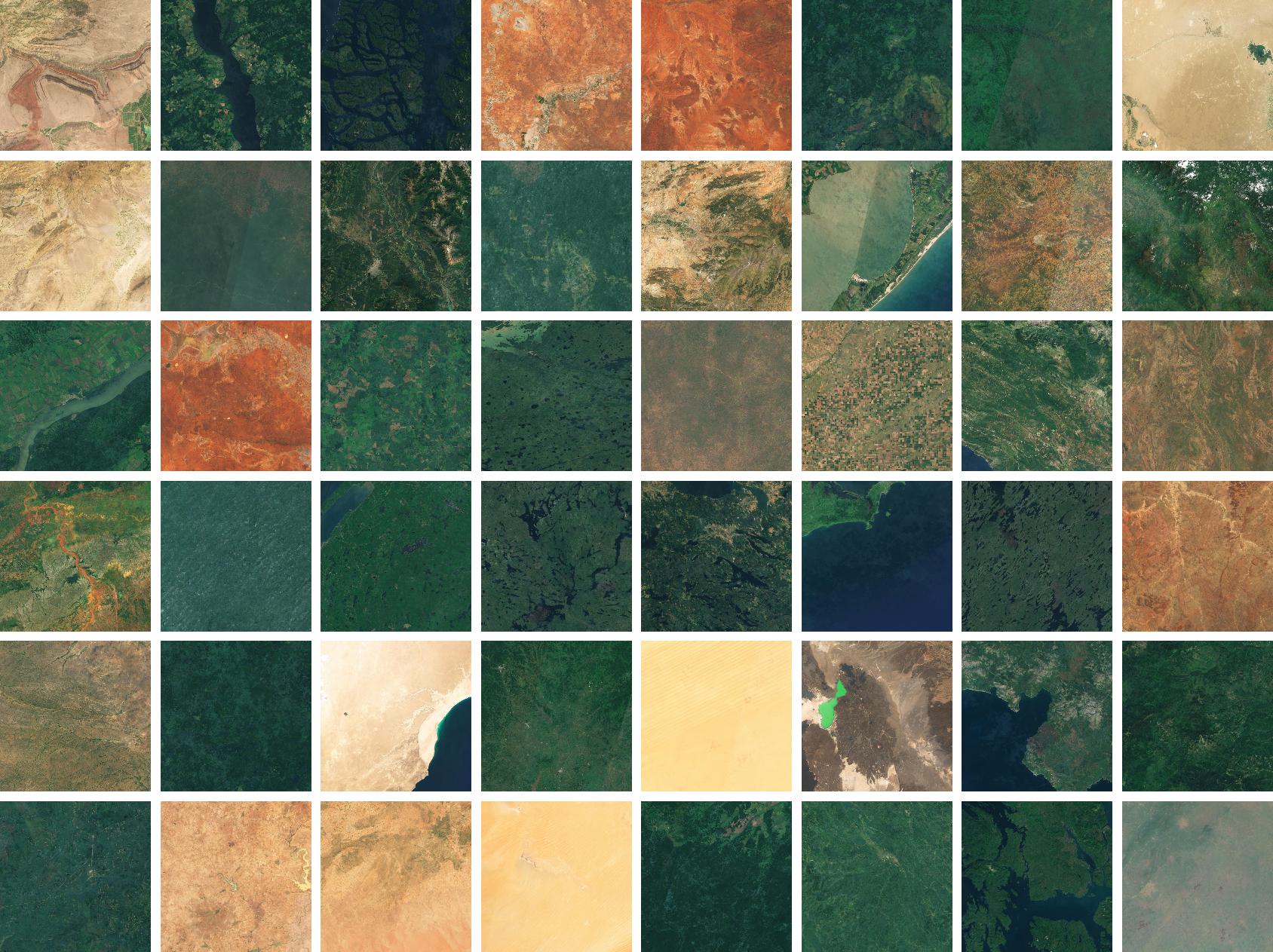}
    \caption{\textbf{Random examples of database images.} All database images come from the EOX Sentinel-2 cloudless collection.\protect\footnotemark 
    }
    \label{fig:supp_db}
    \end{center}
\end{figure*}
\footnotetext{\url{https://s2maps.eu}}

%% file: figures/supp_qualitatives.tex
\begin{figure*}
    \begin{center}
    \includegraphics[width=0.9\linewidth]{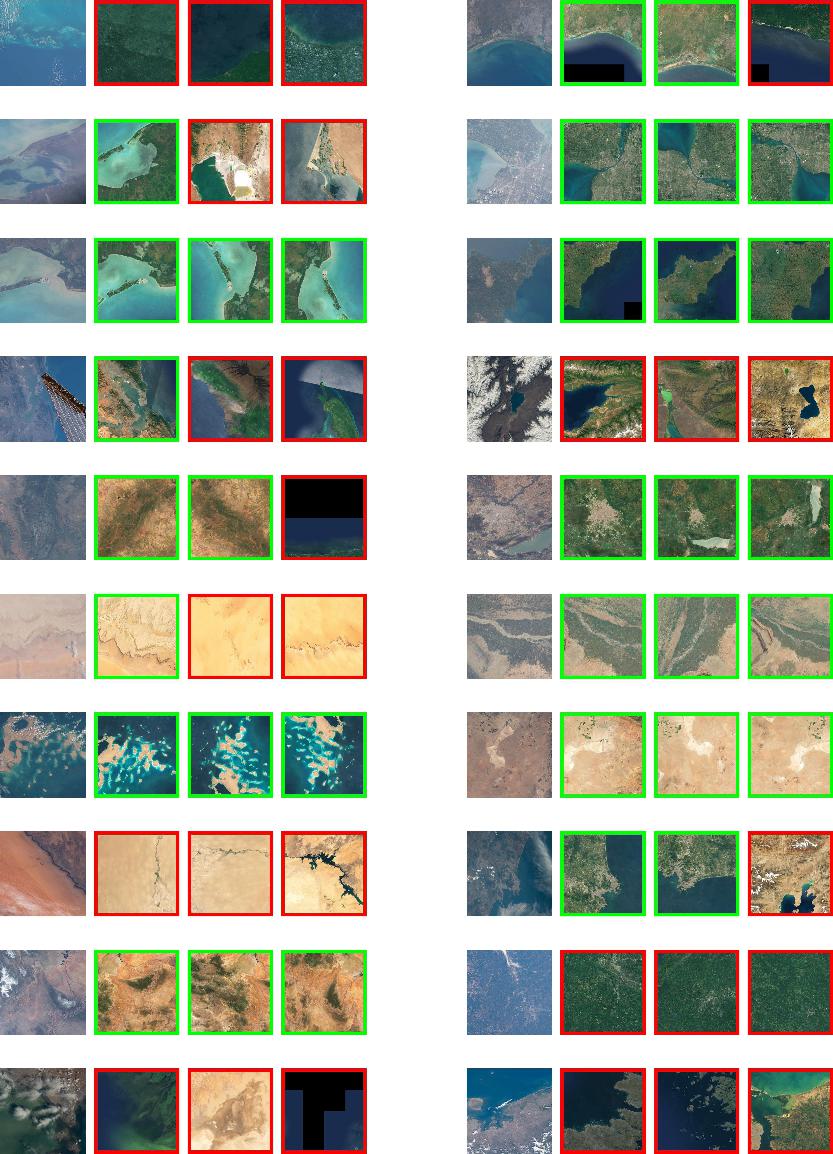}
    \caption{\textbf{Qualitative Examples of EarthLoc Predictions.} Query image and top 3 predictions. Green indicates a correct prediction, red an incorrect prediction. Each half-row is a separate example.}
    \label{fig:supp_qualitatives}
    \end{center}
\end{figure*}

%% file: figures/supp_clusters.tex
\begin{figure*}
    \begin{center}
    \includegraphics[width=0.9\textwidth]{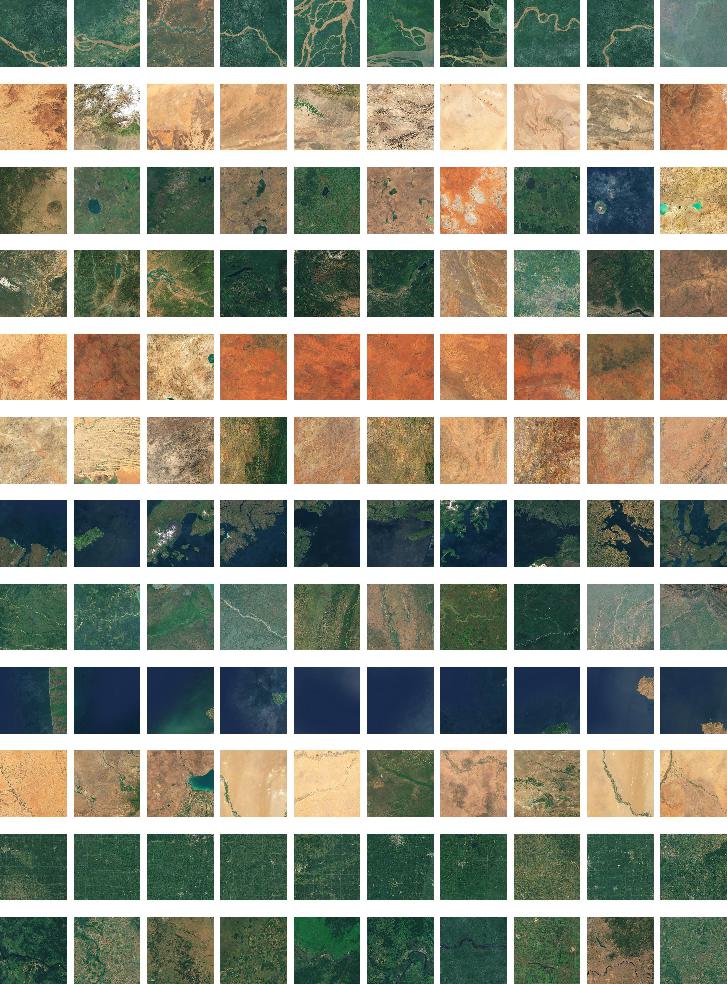}
    \caption{\textbf{Examples of clusters, one cluster per row.} Clusters are formed from images with similar representations. Training batches are selected from images within one cluster (see  \cref{sec:clustered_batches}).
    }
    \label{fig:supp_clusters}
    \end{center}
\end{figure*}

%% file: figures/supp_yw_augm.tex
\begin{figure*}
    \begin{center}
    \includegraphics[width=0.9\textwidth]{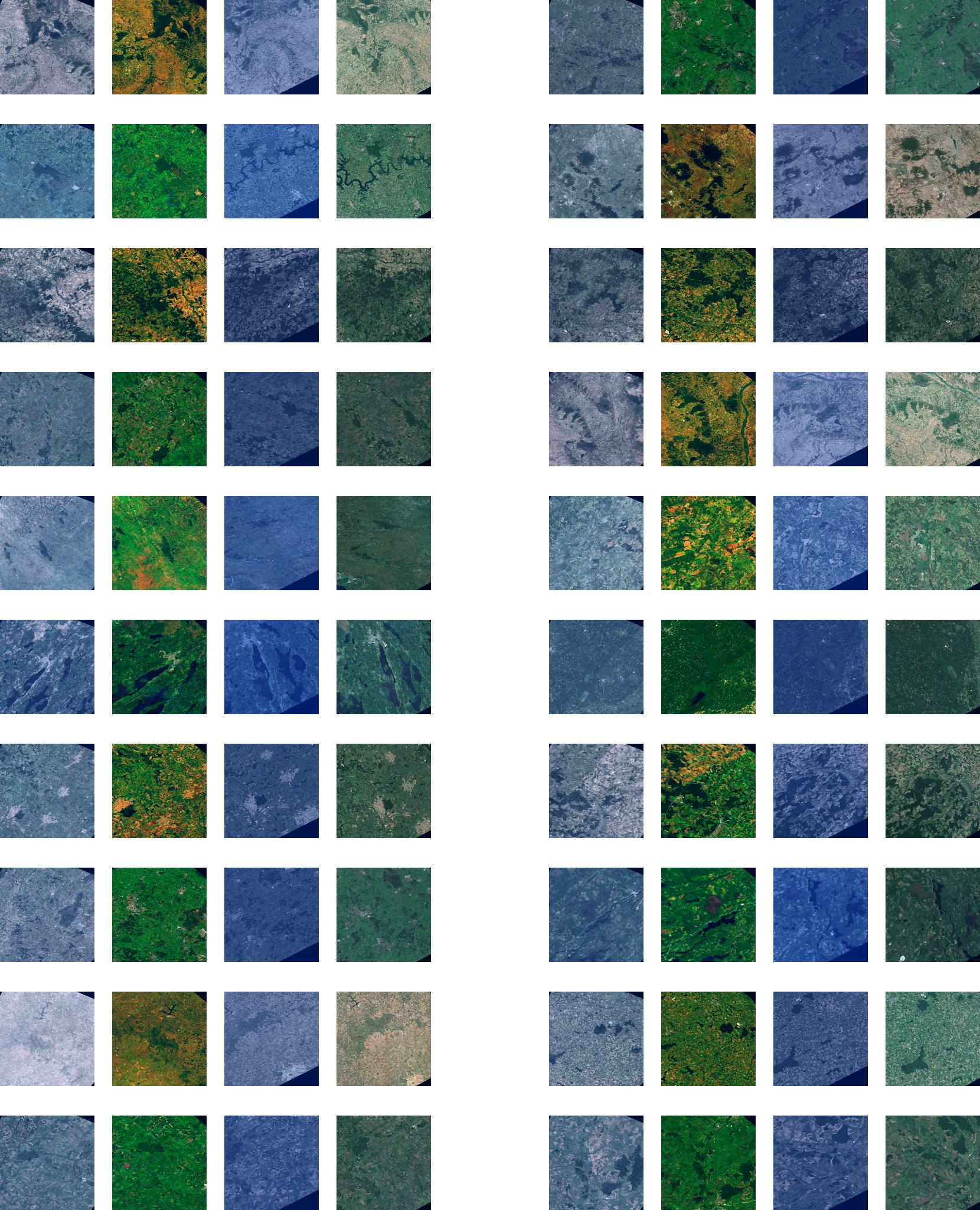}
    \caption{\textbf{An example of a batch}, showing 20 out of 32 quadruplets due to space limits.
    Each half-row of 4 images represents the quadruplet from one \textit{region}. Each \textit{region} has images from 2018, 2019, 2020 and 2021. The same augmentation is applied to all images from a given year, e.g. the images from 2020 receive a blueish color transformation in this batch.
    }
    \label{fig:supp_yw_augm}
    \end{center}
\end{figure*}